\documentclass{article}

\usepackage{microtype}
\usepackage{graphicx}
\usepackage{subfigure}
\usepackage{booktabs} 
\usepackage{xcolor}

\usepackage{hyperref}

\usepackage[accepted]{icml2023}

\usepackage{amsmath}
\usepackage{amssymb}
\usepackage{mathtools}
\usepackage{amsthm}

\usepackage[capitalize,noabbrev]{cleveref}

\theoremstyle{plain}
\newtheorem{theorem}{Theorem}[section]
\newtheorem{proposition}[theorem]{Proposition}

\theoremstyle{definition}

\theoremstyle{remark}

\usepackage[textsize=tiny]{todonotes}

\icmltitlerunning{Probabilistic Contrastive Learning Recovers the Correct Aleatoric Uncertainty}

\begin{document}

\twocolumn[
\icmltitle{Probabilistic Contrastive Learning Recovers \\ the Correct Aleatoric Uncertainty of Ambiguous Inputs}



\icmlsetsymbol{equal}{*}

\begin{icmlauthorlist}
\icmlauthor{Michael Kirchhof}{tub}
\icmlauthor{Enkelejda Kasneci}{muc}
\icmlauthor{Seong Joon Oh}{tub,ai}
\end{icmlauthorlist}

\icmlaffiliation{tub}{University of Tübingen, Germany}
\icmlaffiliation{muc}{TUM University, Munich, Germany}
\icmlaffiliation{ai}{Tübingen AI Center}

\icmlcorrespondingauthor{Michael Kirchhof}{michael dot kirchhof at uni dash tuebingen dot de}

\icmlkeywords{Contrastive Learning, Probabilistic Embeddings, Aleatoric Uncertainty, Heterscedasticity, InfoNCE, Monte-Carlo, Nonlinear ICA, Probabilistic Machine Learning}

\vskip 0.3in
]



\printAffiliationsAndNotice{}  

\begin{abstract}
Contrastively trained encoders have recently been proven to invert the data-generating process: they encode each input, e.g., an image, into the true latent vector that generated the image \cite{dml_inversion}. However, real-world observations often have inherent ambiguities. For instance, images may be blurred or only show a 2D view of a 3D object, so multiple latents could have generated them. This makes the true posterior for the latent vector probabilistic with heteroscedastic uncertainty. 
In this setup, we extend the common InfoNCE objective and encoders to predict latent distributions instead of points. 
We prove that these distributions recover the correct posteriors of the data-generating process, including its level of aleatoric uncertainty, up to a rotation of the latent space. In addition to providing calibrated uncertainty estimates, these posteriors allow the computation of credible intervals in image retrieval. They comprise images with the same latent as a given query, subject to its uncertainty. Code is at \url{https://github.com/mkirchhof/Probabilistic_Contrastive_Learning}.
\end{abstract}

\section{Introduction}
Contrastive learning \cite{chen2020simple} trains encoders to output embeddings that are close to one another for semantically similar inputs and far apart for unsimilar inputs. This general notion of similarity allows transferring pretrained encoders to downstream tasks \cite{wang2022revisiting,ardeshir2022uncertainty,islam2021broad,khosla2020supervised}.

Recently, \citet{dml_inversion} corroborated this intuition by a theoretical result: under weak assumptions, the embeddings learned under an InfoNCE \cite{oord2018representation} loss are exactly equal to the true latent vectors, up to a rotation of the spherical latent space. This comes from a nonlinear Independent Component Analysis (ICA) perspective \cite{comon2010handbook}. It assumes an unknown nonlinear generative process that transforms true latents into our observations. Contrastively trained encoders \emph{invert} this nonlinear function and recover the original latent space, up to a rotation.

\definecolor{blue}{HTML}{4878D0}
\definecolor{red}{HTML}{D65F5F}
\definecolor{green}{HTML}{6ACC64}
\definecolor{orange}{HTML}{EE854A}
\definecolor{grey}{HTML}{797979}
\begin{figure*}[t]
    \centering
    \includegraphics[width=0.9\textwidth]{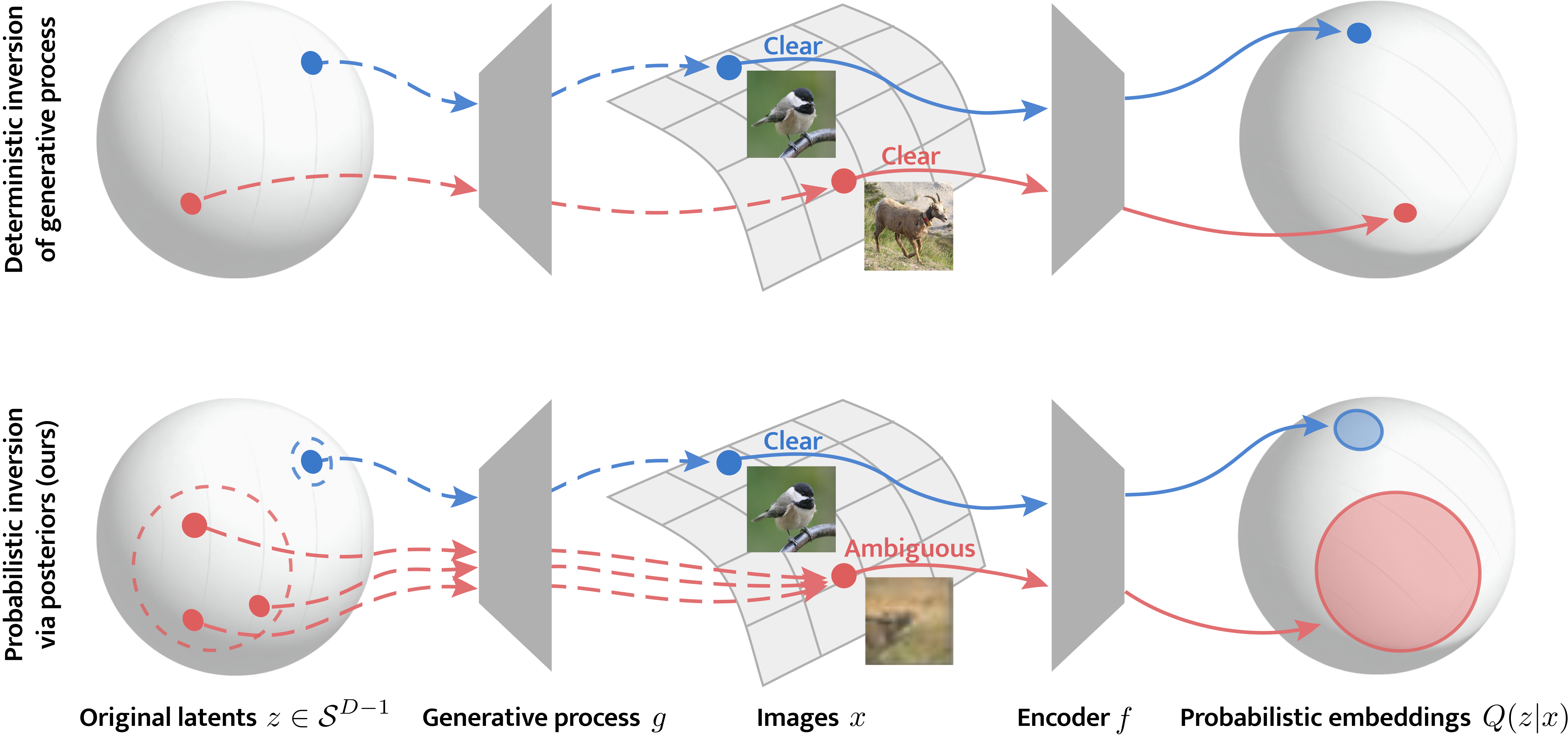}
    \caption{Deteministic encoders embed images to points in the latent space. This recovers the latent vectors that generated them (dashed), up to a rotation (top). However, if an image is ambiguous there are multiple possible latents that could have generated it (bottom). An encoder trained with MCInfoNCE correctly recovers this posterior of the generative process, up to a rotation, from contrastive supervision.}
    \label{fig:overview}
\end{figure*}

This holds for the class of generative processes that are deterministic and injective, so that each image could have been generated by only one latent vector. This is often violated in practice. In \cref{fig:overview}, the lower image of an animal is in low-resolution, so it is impossible to tell which exact species, i.e., which latent variables, underlie the image. In fact, most scenarios in the wild involve some form of such aleatoric uncertainty, including 3D-to-2D projections \cite{chen2021monorun}, partially covered objects \cite{kraus2019uncertainty}, or images with a low resolution or bad crop \cite{li2021spherical}. It also manifests itself outside the image domain, such as in the inherent ambiguity of natural language \cite{chun2022eccv_caption} or measurement noise in general \cite{meech2021algorithm}. Quantifying such uncertainties is a key goal of the recent reliable machine learning efforts \cite{tran2022plex,galil2023what}. This has use cases in safety-critical downstream applications like medical imaging \cite{barbano2022uncertainty}. If an image is too ambiguous, a model can reject it or defer the prediction to a human. Another application is active learning, where we want to choose samples with high uncertainty\cite{lewis1994heterogeneous}. 

This work generalizes the previous theoretical result to this more challenging setting. We do not assume that generative process is an injective and deterministic function, but allow it to be a conditional distribution. We propose Monte-Carlo InfoNCE (MCInfoNCE), a probabilistic analog of InfoNCE. It trains encoders to predict distributions over the possible latents, called probabilistic embeddings \cite{oh2018modeling,shi2019probabilistic}. We prove that MCInfoNCE attains its global minimum when the encoder recovers \textit{the true posteriors} of the generative process, up to a rotation of the latent space; both in terms of both the mean (which latent is most likely to have generated the image) and the variance (the level of aleatoric uncertainty of the individual image). Our work thus generalizes the previous theoretical result in nonlinear ICA to a broader class of generative processes, and provides a theoretical foundation for probabilistic embeddings.

We show empirically that an encoder trained with MCInfoNCE learns the correct posteriors in a controlled experiment with known posteriors. We find that it even provides sensible embeddings when the distribution family or the encoder dimensionality is misspecified and when the generative process may be injective, making it robust in practice. We then show that these predicted uncertainties are consistent with human annotator disagreements reported in the recent CIFAR-10H dataset \cite{peterson2019human}, providing a way to handle uncertainty for high-dimensional inputs. We also demonstrate that knowing the true posteriors enables new applications, such as computing credible intervals for image retrieval tasks. They visualize how uncertain we are about a query image by showing other images that represent the region of latents the query is in with a given probability.

In summary, \textbf{(1)} We extend nonlinear ICA to non-injective non-deterministic generative processes to model realistic input ambiguities. \textbf{(2)} We propose MCInfoNCE for training encoders that predict probabilistic embeddings. \textbf{(3)} We show theoretically and empirically that the predicted posteriors are correct and reflect the true amount of aleatoric uncertainty.

\section{Related Works}

Our work serves as a bridge between the theoretical understanding of contrastive learning via nonlinear ICA, probabilistic embeddings, and recent discussions on the aleatoric uncertainty inherent in vision problems. Below, we discuss how our work extends and connects recent work in these three fields. Extended literature reviews can be found in  \citet{kendall2017uncertainties} and \citet{karpukhin2022probabilistic}.

\textbf{Nonlinear ICA.} From a nonlinear Independent Component Analysis (ICA) perspective \cite{hyvarinen2000independent,comon2010handbook}, images $x$ are generated from ground-truth latent components $z$ via an unknown nonlinear generative process. The goal is to invert it to recover the original latents $z$, which are useful for downstream tasks. This formalization allows for theoretical proofs of which (contrastive) losses achieve this. Building on \citet{wang2020understanding}, \citet{dml_inversion} recently proved that optimizing a contrastive InfoNCE loss \cite{oord2018representation} recovers $z$ up to a rotation of the latent space, as visualized in \cref{fig:overview}. This requires certain assumptions about the generative process. A recent strain of literature seeks to reduce these assumptions \cite{leemann2022disentangling} to allow modeling broader classes of generative processes, bringing the theoretical results closer to practice. Our work broadens this class by no longer requiring the injectivity assumption of \citet{dml_inversion} and at the same time allowing stochasticity. This is made possible by modeling the generative process as a conditional distribution $P(x|z)$ instead of a function, which generalizes the class of generative processes. In the vein of \citet{dml_inversion}, we prove that our contrastive MCInfoNCE loss recovers the correct posterior distribution $P(z|x)$ of the original latents, up to a rotation of the latent space. 

\textbf{Aleatoric Uncertainty.} 
The above generalization allows us to model scenarios in which we encounter aleatoric uncertainty, i.e., the input has reduced information such that $z$ is only recoverable only up to some uncertainty. A prominent practical example is face recognition, where images may be blurred or in low-resolution \cite{shi2019probabilistic,schlett2022face}. Other problems with ambiguous inputs include 3D reconstruction from 2D data \cite{chen2021monorun}, partially occluded traffic participants \cite{kraus2019uncertainty}, or noisy physical sensors \cite{meech2021algorithm}. Such problems with aleatoric uncertainty can be detected by label noise: CIFAR-10H \cite{peterson2019human} comprises multiple labels for each image in the CIFAR-10 test-set, and shows that the more ambiguous an image is, the more annotator labels disagree. This finding occurs in several other recent classification datasets \cite{schmarje2022one,histo,tran2022plex}, but also in more complex tasks such as multimodal visual question answering (VQA). \citet{chun2022eccv_caption} show that there are many possible textual answers to the same visual prompt because language is inherently more ambiguous than vision; i.e., language has more aleatoric uncertainty. Our MCInfoNCE loss explicitly accounts for these uncertainties and learns the correct level of aleatoric uncertainty, which we demonstrate on high-dimensional image inputs. 

\textbf{Probabilistic Embeddings.} An emerging approach to modeling this uncertainty is to have encoders predict distributions over the latent space instead of point estimates. There are three main lines of work to learn these probabilistic embeddings. The first idea is to compute a match probability between point estimates, but to integrate it over the predicted distributions. This idea was pioneered via Hedged Instance Embeddings (HIB) \cite{oh2018modeling} and has since been successfully extended, e.g., to the above multimodal VQA  problem \cite{chun2021probabilistic,neculai2022probabilistic}. A second line of works turns existing losses into probabilistic ones by integrating the whole loss over the predicted probabilistic embeddings \cite{scott2021mises,roads2021enriching}. Our MCInfoNCE extension of InfoNCE demonstrates that this blueprint strategy can inherit the properties of the original losses, like \citet{dml_inversion}'s identifiability theorem. The third line of works provides distribution-to-distribution distances to replace point-to-point distances in losses. The most popular approach is the expected likelihood kernel (ELK) \cite{jebara2003bhattacharyya,shi2019probabilistic}. It has recently shown success even in high dimensional embedding spaces \cite{kirchhof2022non,karpukhin2022probabilistic}. Yet, there is no answer to whether and in what sense the predicted probabilistic embeddings, and in particular their variances, are \textit{correct}. Our work answers this question through its proof and a controlled experiment where the true posteriors are recovered. 
The experiments on CIFAR-10H further ground this theoretical correctness in the human perception of uncertainty. We also show novel practical applications of probabilistic embeddings, such as retrieving credible intervals on which latents the image might show. 

\section{Probabilistic Generative Processes}

In this section, we extend the generative processes commonly used in nonlinear ICA to non-injective, randomized ones. This allows modeling real-world image distributions better and serves as a framework for the upcoming proof.

Let us first understand the class of generative processes for which \citet{dml_inversion} prove identifiability. They take the nonlinear ICA perspective that there is a natural generative process $g$ that transforms latent components $z \in \mathcal{Z}$ into the images $x = g(z)$ we observe, as shown in \cref{fig:overview}. Following the popular cosine-based similarity comparisons \cite{deng2019arcface,teh2020proxynca}, $\mathcal{Z}$ is assumed to be a $D$-dimensional hypersphere $\mathcal{Z} = \mathcal{S}^{D-1}$. We are interested in recovering the latents $z$ that underlie the images $x$, because they are low-dimensional descriptions useful for downstream tasks. To formalize this problem, they assume that $g: \mathcal{Z} \rightarrow \mathcal{X}$ is an injective (and deterministic) function. Thus, only one latent $z$ can correspond to each image $x$, and $g$ is invertible. They prove that an encoder $f$ trained with a contrastive InfoNCE loss achieves this inversion and recovers the correct latent $z$, i.e., $f(x) = f(g(z)) = \hat{z} = R z$, up to an orthogonal rotation $R$ of the learned embedding space.

However, let us move on to setups where an image $x$ may be motion blurred, low-resolution, or partially obscured. For instance, a 2D projection $x$ of a 3D object $z$ does not show the back part of $z$, and there are several possible $z$ that could have generated $x$. In other words, the generative process $g$ is non-injective and the best our encoder can do is to recover the set of possible latents $\{\hat{z}|g(\hat{z}) = x\}$. Further, $g$ may be stochastic. E.g., a random patch of pixels may be occluded, or the image may be zoomed in and show only a random crop of $z$. The best the encoder can do is to predict a posterior over the possible latents, see \cref{fig:overview}.

The common denominator of these setups is that $g$ loses information about $z$ and $x$ becomes ambiguous. To subsume them, we can model $g$ as a likelihood $P(x|z)$. This general formulation allows for a large class of operations within $g$. However, this generality comes at the cost that $P(x|z)$ can be very complicated and difficult to parameterize. We therefore apply a \emph{posterior trick}: instead of explicitly characterizing $g$ by $P(x|z)$ we implicitly characterize it by its posteriors $P(z|x)$.
We parameterize $P(z|x)$ by simple von Mises-Fisher distributions $\text{vMF}(z; \mu(x), \kappa(x))$: 
\begin{align}
P(z|x) = C(\kappa(x)) e^{\kappa(x) \mu(x)^\top z}\,\,.
\end{align}
This distribution on $\mathcal{S}^{D-1}$ is unimodal around the location parameter $\mu(x) \in \mathcal{Z}$ with a certain concentration (i.e., an \emph{inverse} variance) $\kappa(x) \in \mathbb{R}_{> 0}$, and a normalizing constant $C(\cdot)$. The functions $\mu: \mathcal{X} \rightarrow \mathcal{S}^{D-1}$ and $\kappa: \mathcal{X} \rightarrow \mathbb{R}_{> 0}$ fully parameterize the posterior of each image $x$. In particular, $\kappa(\cdot)$ represents the aleatoric uncertainty due to information loss, which can be heterogeneous across the images.

The intuition behind modeling the posterior of the generative process as a vMF is that latents of degraded images can usually be located down to sets of semantically similar rather than very dissimilar latents. This is reflected in the unimodality of the vMF and its use of the dot product, which commonly represents how semantically similar two latents are. There may still be images where it is impossible to tell which highly dissimilar latents they show. In these cases, $\kappa(x)$ is low and the posterior spreads broadly across the latent space. At the other end of the spectrum, as $\kappa(x) \rightarrow \infty$, $P(z|x)$ converges to a Dirac distribution. This allows modelling deterministic and injective generative processes as in \citet{dml_inversion}. This makes the vMF a reasonable and flexible choice for the posterior of generative processes.

\section{Probabilistic Contrastive Learning}
This section presents our main theoretical result: a probabilistic encoder trained under an MCInfoNCE loss recovers the true posteriors of probabilistic generative processes, up to a rotation, from simple contrastive supervision.

\subsection{\resizebox{0.43\textwidth}{!}{MCInfoNCE for Probabilistic Contrastive Learning} \label{sec:setup}}
Let us first formalize the contrastive learning setup. Each training triplet comprises a reference sample $x$ along with a positive (similar) sample $x^+$ and negative (dissimilar) samples $x^-_1, \dotsc, x^-_M$ against which it is to be contrasted. 
As introduced in the previous section, we assume that these samples are generated from corresponding latents $z, z^+, z^-_1, \dotsc, z^-_M$. Following \citet{dml_inversion}, the reference $z$ is drawn from the marginal distribution in the latent space, a uniform distribution. The positive sample $z^+$ is drawn from a close region around $z$, while negatives $z^-_1, \dotsc, z^-_M$ are random i.i.d. draws from the marginal:
\begin{align}
    z &\sim P(z) = \text{Unif}(z;\mathcal{S}^{D-1}), \label{form:gen1}\\
    z^+ &\sim P(z^+|z) = \text{vMF}(z^+;z, \kappa_\text{pos}), \label{form:gen2}\\
    z^-_m &\sim P(z^- | z) =: P(z^-) = \text{Unif}(z^-;\mathcal{S}^{D-1}). \label{form:gen3}
\end{align}
The fixed constant $\kappa_\text{pos} > 0$ controls how close latents must be to be considered positive to each other\footnote{$\kappa_\text{pos}$ should not to be confused with $\kappa(x)$, which controls the heteroscedastic uncertainty of the generative process.}. This formalization of contrastive learning ensures that positive samples are semantically similar and negatives are dissimilar. \citet{dml_inversion} showed this is the generative process InfoNCE implicitly assumes. The probabilistic generative process comes into play when the latents $z, z^+, z^-_1, \dotsc, z^-_M$ are transformed into observations $x, x^+, x^-_1, \dotsc, x^-_M$ via $P(x|z)$. This defines $P(x)$, $P(x^+|x)$, and $P(x^-)$, and thus our contrastive training data $(x, x^+, x^-_1, \dotsc, x^-_M)$.

Our Monte-Carlo InfoNCE (MCInfoNCE) loss is
\begin{align}
    L_f \hspace{-0.5mm}:= \hspace{-0.5mm} & -\log \hspace{-1.45cm} \mathop{\mathbb{E}}\limits_{\substack{z \sim Q(z|x) \\ z^+ \sim Q(z^+|x^+) \\ z^-_m \sim Q(z^-_m|x^-_m), m=1, \dotsc, M}} \hspace{-0.9cm} \left( \hspace{-1mm} \frac{e^{\kappa_\text{pos} z^\top z^+}}{\frac{1}{M} e^{\kappa_\text{pos} z^\top z^+} \hspace{-1mm} + \hspace{-0.5mm}\frac{1}{M} \hspace{-1mm} \sum\limits_{m=1}^M e^{\kappa_\text{pos} z^\top z^-_m}} \hspace{-1mm} \right) \hspace{-1mm}\label{form:loss}
\end{align}
and is evaluated over the contrastive training dataset via
\begin{align}
    \mathcal{L} :=& \hspace{-0.34cm} \mathop{\mathbb{E}}\limits_{\substack{x \sim P(x) \\ x^+ \sim P(x^+|x) \\ x^-_m \sim P(x^-), m=1, \dotsc, M}} \hspace{-1cm} \left( L_f\left(x, x^+, \{x_m^-\}_{m=1, \dotsc, M}\right)\right) \hspace{-0.5mm}.
\end{align}
This probabilistically generalizes the widely used InfoNCE family \cite{oord2018representation}, and, in the limit of $M \rightarrow \infty$, SimCLR \cite{chen2020simple}. Instead of outputting a point embedding, the encoder $f$ we train outputs probabilistic embeddings $Q(z|x) := \text{vMF}(z; \hat{\mu}(x), \hat{\kappa}(x))$ by predicting $f(x) = (\hat{\mu}(x), \hat{\kappa}(x))$. The InfoNCE fraction within $L_f$ is evaluated over these posteriors. In practice, we backpropagate through $K=512$ MC samples via a reparametrization trick for vMFs \cite{s-vae18,ulrich1984computer}:
\begin{align}
    L_f \hspace{-1mm}\approx \hspace{-1mm} -\log \hspace{-1mm}\left( \hspace{-1mm}\frac{1}{K} \hspace{-0.5mm}\sum\limits_{k=1}^K \frac{e^{\kappa_\text{pos} z_k^\top z_k^+}}{\frac{1}{M} e^{\kappa_\text{pos} z_k^\top z_k^+} \hspace{-1mm} + \hspace{-0.5mm}\frac{1}{M} \hspace{-1mm} \sum\limits_{m=1}^M e^{\kappa_\text{pos} z_k^\top z^-_{m, k}}} \hspace{-1mm}  \right)\hspace{-1mm}. \label{form:MC}
\end{align}
The only training data for MCInfoNCE are contrastive examples, without any additional supervision on the true aleatoric uncertainty $\kappa(x)$ or the generative latents $z$.

\subsection{Provably Learning the Correct Posteriors}

We prove below that the optimizer of this loss learns the \textit{correct} latent posteriors. More precisely, it predicts the correct location $\hat{\mu}(x) = R \cdot \mu(x)$, up to a constant orthogonal rotation $R$ of the latent space, and the correct level of ambiguity $\hat{\kappa}(x) = \kappa(x)$ for each observation $x$. To prove this, we first show that MCInfoNCE is a cross-entropy between the generative process and the learned contrastive encoder (Proposition~\ref{form:equal_marginals}). This means that the loss matches the expected positivity of a pair $(x, x^+)$ computed using the true $P(z|x)$ to that computed using $Q(z|x)$. We then show that this expected positivity can be written as a function and depends only on $(\mu(\cdot)^\top\mu(\cdot), \kappa(\cdot))$, resp. $(\hat{\mu}(\cdot)^\top\hat{\mu}(\cdot), \hat{\kappa}(\cdot))$ (Proposition~\ref{eq:mono}). Due to monotonicity, the predicted function value can only match that of the generative process if their arguments $(\mu(\cdot)^\top\mu(\cdot), \kappa(\cdot))$ and $(\hat{\mu}(\cdot)^\top\hat{\mu}(\cdot), \hat{\kappa}(\cdot))$ are equal (\cref{eq:equalkappa}). In summary, the posteriors must be equal, up to a rotation of the latent space (Theorem~\ref{eq:final}).

First, we generalize \citet{dml_inversion} and \citet{wang2020understanding} to probabilistic generative processes.
\begin{proposition}[$\mathcal{L}$ is minimized iff expected positivity matches] \label{form:equal_marginals}
    Let the latent marginal $P(z) = \int P(z|x) dP(x)$ and $\int Q(z|x) dP(x)$ be uniform. $\lim_{M \rightarrow \infty}\mathcal{L}$ attains its minimum when $\forall x, x^+ \in \{x \in \mathcal{X} | P(x) > 0\}$
    \begin{align*}
       & \iint Q(z|x) Q(z^+|x^+) P(z^+|z) dz^+ dz = \\
       & \iint P(z|x) P(z^+|x^+) P(z^+|z) dz^+ dz \,\,.
    \end{align*}
\end{proposition}

The intuition is that MCInfoNCE corresponds to a cross-entropy between the true latents and our model predictions. This characterizes the solution set: An encoder $Q$ minimizes MCInfoNCE if and only if the chance of $(x, x^+)$ being a positive pair \textit{computed using $Q$} is equal to the \textit{true chance} of being a positive pair \textit{computed using the GT distribution $P$} for all data pairs $(x, x^+)$. We refer to this chance, the upper integral, as expected positivity. Next, we prove that the equality of the expected positivities implies that the predicted posteriors $Q$ must be equal to the GT $P$, up to the mentioned rotations. To this end, we first find that the expected positivity marginalizes out all random variables and can be written as a \textit{function} of $\mu(x)$ and $\kappa(x)$.

\begin{proposition}[Expected positivity is a function] \label{eq:mono}
    Let $P(z|x)$ and $P(z^+|z)$ be vMF distributions as defined in \cref{sec:setup}. Given $x, x^+ \in \mathcal{X}$, we can rewrite \begin{align}
        & \iint P(z|x) P(z^+|x^+) P(z^+|z) dz^+ dz \\
        & =: h_{\kappa_\text{pos}}(\mu(x)^\top \mu(x^+), \kappa(x), \kappa(x^+)), 
    \end{align} 
    i.e., as a function $h_{\kappa_\text{pos}}$ that depends only on $\mu(x)^\top \mu(x^+), \kappa(x),$ and $\kappa(x^+)$. The same function can be used for $\hat{\mu}(x)^\top \hat{\mu}(x^+), \hat{\kappa}(x), \hat{\kappa}(x^+)$:
    \begin{align}
        & \iint Q(z|x) Q(z^+|x^+) P(z^+|z) dz^+ dz \\
        & = h_{\kappa_\text{pos}}(\hat{\mu}(x)^\top \hat{\mu}(x^+), \hat{\kappa}(x), \hat{\kappa}(x^+)). 
    \end{align}
\end{proposition}

The key is that the expected positivities calculated using $Q$ and $P$ have the \textit{same} functional form $h_\text{pos}$; they differ only in their arguments, where they use either the true $\kappa(x), \mu(x)$ or the predicted $\hat{\kappa}(x), \hat{\mu}(x)$. What remains to show is that the expected positivities can only be equal if the arguments match, i.e.,  $\hat{\kappa}(x) = \kappa(x)$ and $\hat{\mu}(x)^\top \hat{\mu}(x^+) = \mu(x)^\top \mu(x^+)$. \cref{eq:equalkappa} proves this via some monotonicities of $h_\text{pos}$.

\begin{proposition}[Arguments of $h_\text{pos}$ must be equal] \label{eq:equalkappa}
    Define $h_\text{pos}$ as in \cref{eq:mono}. Let $\mathcal{X}' \subseteq \mathcal{X}$, $\mu, \hat{\mu}: \mathcal{X}' \rightarrow \mathcal{Z}$, $\kappa, \hat{\kappa}: \mathcal{X}' \rightarrow \mathbb{R}_{> 0}$, $\kappa_\text{pos} > 0$. If $h_{\kappa_\text{pos}}(\hat{\mu}(x)^\top \hat{\mu}(x^+), \hat{\kappa}(x), \hat{\kappa}(x^+)) = h_{\kappa_\text{pos}}(\mu(x)^\top \mu(x^+), \kappa(x), \kappa(x^+))$ $\forall x, x^+ \in \mathcal{X}'$, then 
    \begin{align}
        & \hat{\mu}(x)^\top \hat{\mu}(x^+) = \mu(x)^\top \mu(x^+) \text{ and} \label{eq:mazur} \\
        & \hat{\kappa}(x) = \kappa(x) \,\,\, \forall x, x^+ \in \mathcal{X}'.
    \end{align}
\end{proposition}

In the above \cref{eq:mazur}, the pairwise cosine similarities in the true and the predicted latent space can only be equal if the two spaces are the same up to a rotation, i.e., $\hat{\mu}(x) = R \mu(x)$. This is ensured by the Extended Mazur-Ulam Theorem \citep{dml_inversion}.
We can now combine these ingredients to derive our main result: If an encoder minimizes the MCInfoNCE loss, then it must have identified the correct posteriors, up to a constant orthogonal rotation of the latent space. 

\begin{theorem}[$\mathcal{L}$ identifies the correct posteriors] \label{eq:final}
Let $\mathcal{Z} = \mathcal{S}^{D-1}$ and $P(z) = \int P(z|x) dP(x)$ and $\int Q(z|x) dP(x)$ be the $\text{Unif}(z;\mathcal{Z})$. Let $g$ be a probabilistic generative process defined in Formulas \ref{form:gen1}, \ref{form:gen2}, and \ref{form:gen3} with known\footnote{In practice, $\kappa_\text{pos}$ is a tuneable temperature hyperparameter.} $\kappa_\text{pos}$. Let $g$ have vMF posteriors $P(z|x) = \text{vMF}(z;\mu(x), \kappa(x))$ with $\mu: \mathcal{X} \rightarrow \mathcal{S}^{D-1}$ and $\kappa: \mathcal{X} \rightarrow \mathbb{R}_{> 0}$. Let an encoder $f(x)$ parametrize vMF distributions $\text{vMF}(z; \hat{\mu}(x), \hat{\kappa}(x))$. Then $f^* = \arg\min_f \lim_{M \rightarrow \infty} \mathcal{L}$ has the correct posteriors up to a rotation, i.e., $\hat{\mu}(x) = R \mu(x)$ and $\hat{\kappa}(x) = \kappa(x)$, where $R$ is an orthogonal matrix, $\forall x \in \{x \in \mathcal{X} | P(x) > 0\}$.
\end{theorem}

This generalizes the recent results of \citet{dml_inversion} to the broader family of probabilistic generative processes. MCInfoNCE recovers not only the correct (mean) embeddings $\mu(x)$ under a noisy and non-injectivity generator, but also the heterogeneous aleatoric uncertainty $\kappa(x)$. 

\section{Experiments}

\subsection{MCInfoNCE Learns the Correct Posteriors} \label{sec:toy}

In this section, we experimentally confirm the theoretical result that \textit{probabilistic embeddings learned under a MCInfoNCE loss recover the correct posteriors up to a rotation}. We also test its robustness to violated assumptions.

\textbf{Setup.} To test whether MCInfoNCE recovers the correct posteriors, we need a controlled experiment where the true posteriors of the generative process are known. Previous nonlinear ICA experiments randomly initialize a multi-layer perceptron (MLP) as the nonlinear data-generating process and train a second one to invert it \cite{hyvarinen2017nonlinear,dml_inversion}. In our probabilistic setup we randomly initialize two MLPs to parameterize $\mu(x)$ and $\kappa(x)$ of the vMF posteriors of the generative process. The MLP for $\mu(x)$ outputs normalized vectors of dimension $D=10$ and the MLP for $\kappa(x)$ outputs a scalar $\tilde{\kappa}(x)$ wrapped in an exponential Softplus function $\kappa(x) = 1 + \exp(\tilde{\kappa}(x))$ to ensure the strict positivity of $\kappa(x)$ \cite{li2021spherical,shi2019probabilistic}.
We sample contrastive training data $(x, x^+, (x^-_m)_{m=1, \dotsc, M})$ from the generative process parameterized by $\mu(x)$ and $\kappa(x)$ via rejection sampling, as explained in the supplementary.
On this data, we train two MLPs to predict $\hat{\mu}(x)$ and $\hat{\kappa}(x)$. All hyperparameters of the generative process and MLP architectures follow the deterministic counterpart of this experiment in \citet{dml_inversion} and are reported in the supplementary. 

\definecolor{lightgrey}{HTML}{adb3b7}
\newcommand{\res}[2]{$#1 \color{lightgrey} \pm \small#2$}
\newcommand{\bres}[2]{$\bm{#1} \pm \bm{#2}$}
\begin{table}[t]
    \centering
    \caption{MCInfoNCE recovers the generative processes' true posteriors for various degrees of ambiguity and even in the limit of an injective generative process. Mean $\pm$ std. err. for five seeds.}
    \label{tab:toy_correctsetups}
    \resizebox{0.48\textwidth}{!}{
    \begin{tabular}{lcccc}
    \toprule
          &  \multicolumn{2}{c}{True vs Pred. Location $\hat{\mu}(x)$} & \multicolumn{2}{c}{True vs Pred. Certainty $\hat{\kappa}(x)$}\\
         Generative Process Ambiguity & RMSE $\downarrow$ & Rank Corr. $\uparrow$ & RMSE $\downarrow$ & Rank Corr. $\uparrow$ \\
         \midrule
         Ambiguous ($\kappa(x) \in [16, 32]$) & \res{0.04}{0.00} & \res{0.99}{0.00} & \res{6.15}{0.61} & \res{0.82}{0.04} \\
         Clear ($\kappa(x) \in [64, 128]$) & \res{0.05}{0.00} & \res{0.98}{0.00} & \res{125.02}{10.64}& \res{0.64}{0.04} \\
         Injective ($\kappa(x) = \infty$) & \res{0.05}{0.01} & \res{0.98}{0.00} & \multicolumn{2}{c}{$\hat{\kappa}(x) \rightarrow \infty$} \\
         \bottomrule
    \end{tabular}
    }
\end{table}

\begin{figure}[t]
    \centering
    \includegraphics[width=0.23\textwidth, trim=1.35cm 2.16cm 2cm 1.75cm, clip]{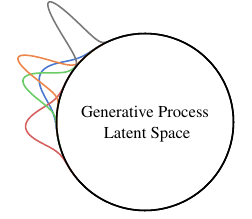} \includegraphics[width=0.23\textwidth, trim=2cm 2.31cm 1.35cm 1.75cm, clip]{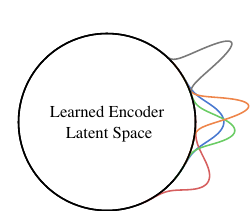}
    \caption{Five posteriors of the generative process and the encoder trained in a run with a 2D latent space. The encoder correctly predicts the posteriors of the generative process, up to a rotation: Rank corr. between $\hat{\mu}(x)$ and the true $\mu(x)$ is \res{1.00}{0.00} (RMSE \res{0.05}{0.00}) and that of $\hat{\kappa}(x)$ is \res{0.82}{0.05} (RMSE \res{2.89}{0.56}).}
    \label{fig:2dplot}
    \vspace{-2mm}
\end{figure}

\textbf{Metrics.} To quantify if the predicted posteriors are correct up to a rotation, i.e., $\hat{\kappa}(x) = \kappa(x)$ and $\hat{\mu}(x) = R \mu(x)$ with an orthogonal matrix $R$, we compare $\hat{\kappa}(x)$ to $\kappa(x)$ on $10^4$ samples of $x$ and compare $\hat{\mu}(x_1)^\top \hat{\mu}(x_2)$ to $\mu(x_1)^\top \mu(x_2)$ on all pairs $(x_1, x_2)$ of the $10^4$ samples. We use the root mean square error (RMSE) to test for exact correctness and Spearman's rank correlation (Rank Corr.) to test for correct ordering. The latter is sufficient in practical scenarios that are invariant to scale, such as retrieval based on embedding distances $\hat{\mu}(x_1)^\top \hat{\mu}(x_2)$ or abstention from prediction based on a threshold of the predicted certainty $\hat{\kappa}(x)$. 

\begin{figure}[t]
    \centering
    \includegraphics{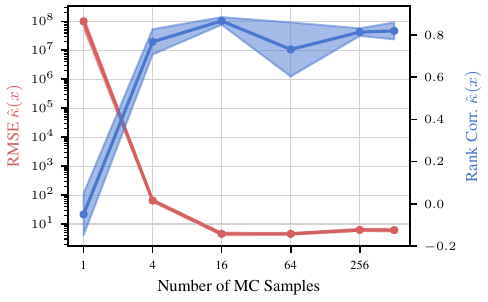}
    \caption{The marginal likelihood approximation bias diminishes with sufficient MC samples. Mean $\pm$ std. err. for five seeds.}
    \label{fig:mc}
\end{figure}

\textbf{Results.} \cref{tab:toy_correctsetups} shows that MCInfoNCE recovers the correct posteriors of ambiguous inputs up to a high rank correlation of $0.99$ for $\hat{\mu}(x)$ and $0.82$ for $\hat{\kappa}(x)$. \cref{fig:2dplot} visualizes this in a simplified 2D case. The learned latent space equals the true latent space up to a rotation. However, we can see in \cref{tab:toy_correctsetups} that $\hat{\kappa}(x)$ tends to be overconfident (RMSE $=125.02$) especially for high values of $\kappa(x) \in [64, 128]$ (yet, the ranking is still largely preserved, Rank Corr. $=0.64$). This is because Formula \ref{form:MC} is a biased MC estimator of the loss in Formula \ref{form:loss}. This is also known as marginal likelihood estimation problem \cite{perrakis2014use,burda2015importance}. The bias decreases with the number of MC samples, as shown in \cref{fig:mc}. In the standard setup with $\kappa(x) \in [16, 32]$, it is largely mitigated with 16 samples (RMSE $=4.55$), or already with 4 samples if only the relative ordering of the samples matters in practice (Rank Corr. $= 0.77$). This coincides with the range of number of MC samples used by other probabilistic embedding losses: \citet{oh2018modeling} use 10 and \citet{kirchhof2022non} use 5. In summary, MCInfoNCE behaves as theoretically expected and fulfills our main theoretical hypothesis.

\begin{table}[t]
    \centering
    \caption{MCInfoNCE predicts sensible vMF posteriors if the true generative posteriors are non-vMF. Mean $\pm$ std. err. for five seeds.} 
    \label{tab:toy_distrfamily}
    \resizebox{0.45\textwidth}{!}{
    \begin{tabular}{lcccc}
    \toprule
          &  \multicolumn{2}{c}{True vs Pred. Location $\hat{\mu}(x)$} & \multicolumn{2}{c}{True vs Pred. Spread}\\
         Posterior & RMSE $\downarrow$ & Rank Corr. $\uparrow$ & RMSE $\downarrow$ & Rank Corr. $\uparrow$ \\
         \midrule
         vMF & \res{0.04}{0.00} & \res{0.99}{0.00} & \res{0.05}{0.00} & \res{0.75}{0.04} \\
         Gaussian & \res{0.04}{0.00} & \res{0.99}{0.00} & \res{0.04}{0.00} & \res{0.70}{0.05} \\
         Laplace & \res{0.05}{0.01} & \res{0.98}{0.00}& \res{0.02}{0.00} & \res{0.66}{0.06} \\
         \bottomrule
    \end{tabular}
    }
\end{table}

\begin{figure}[t]
    \centering
    \includegraphics{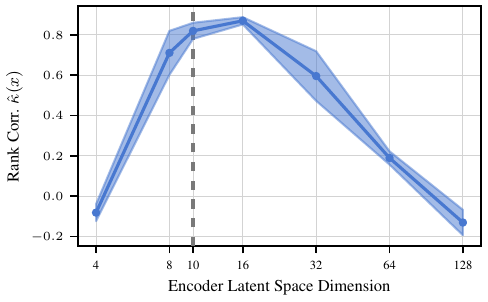}
    \caption{MCInfoNCE learns good $\hat{\kappa}(x)$ even when the encoder latent space dimension mismatches the true generative dimensionality ($D=10$). Mean $\pm$ std. err. for five seeds.}
    \label{fig:encoder_size}
\end{figure}

\textbf{Violated Assumptions.} We test MCInfoNCE in setups where its assumptions are violated. First, we change the posterior of the generative process to Gaussian and Laplace distributions on $\mathcal{S}^{D-1}$ while the encoder still predicts vMFs. Since these distributions have incomparable variance parameters, we measure their spread by the avg. absolute cosine distance from the mode. \cref{tab:toy_distrfamily} shows that the vMFs model Gaussians almost as well as vMFs (Rank Corr. $0.70$ vs $0.75$), since Gaussians with normalized outputs are similar to vMFs \cite{mardia2000directional}. For Laplace, the encoder predicts vMFs with high concentrations ($\hat{\kappa}(x) \approx 2000$), because the Laplace distribution is more concentrated around its mode than the vMF the encoder uses. Second, we over- and underparameterize the latent dimension of the encoder compared to that of the generative process ($D=10$). \cref{fig:encoder_size} shows that encoder dimensions between $8$ and $32$ still all yield $\hat{\kappa}$ predictions with a Rank Corr. $\geq 0.6$. Third, we test the behaviour of MCInfoNCE when the generative process is injective and deterministic, i.e., when all posteriors are Diracs. This is a limiting case of the vMFs the encoder uses. \cref{tab:toy_correctsetups} shows that the predicted vMFs converge to infinite concentrations $\hat{\kappa}(x)$, recovering the Diracs. Last, the uniformity assumption was violated in all experiments as we only ensured $\mu(x)$ to be not collapsed, but not necessarily fully spread around $\mathcal{S}^{D-1}$. In summary, these results indicate that MCInfoNCE is a robust approach even when characteristics of the generative process such as its (non-) injectivity, posterior family, or dimension are unknown.

\begin{table}[t]
    \centering
    \caption{Besides MCInfoNCE, ELK also gives correct probabilistic embeddings. Mean $\pm$ std. err. for five seeds. \vspace{1mm}}
    \label{tab:bench_results}    
    \resizebox{0.48\textwidth}{!}{
    \begin{tabular}{lcccc}
    \toprule
          &  \multicolumn{2}{c}{True vs Pred. Location $\hat{\mu}(x)$} & \multicolumn{2}{c}{True vs Pred. Certainty $\hat{\kappa}(x)$}\\
         Loss & RMSE $\downarrow$ & Rank Corr. $\uparrow$ & RMSE $\downarrow$ & Rank Corr. $\uparrow$ \\
         \midrule
         HIB & \res{0.18}{0.02} & \res{0.82}{0.03} & $10^{14} \color{lightgrey} \pm 10^{14}$ & \res{-0.02}{0.09} \\
         ELK & \res{0.02}{0.00} & \res{1.00}{0.00} & \res{21.70}{0.31} & \res{0.92}{0.00} \\
         MCInfoNCE & \res{0.04}{0.00} & \res{0.99}{0.00} & \res{6.15}{0.61} & \res{0.82}{0.04} \\
         \bottomrule
    \end{tabular}
    }
\end{table}

\textbf{Further losses.} Recent literature has proposed other losses to predict probabilistic embeddings. We investigate their empirical successes further under our experimental setup to find whether they \emph{exactly} match the true posteriors. We reimplement Hedged Instance Embeddings (HIB) \cite{oh2018modeling} and Expected Likelihood Kernels (ELK) \cite{kirchhof2022non} and modify them to our contrastive setup, as detailed in the supplementary. All losses are hyperparameter tuned via grid search. \cref{tab:bench_results} shows that all losses recover $\mu(x)$ with a Rank Corr. $\geq 0.82$ despite the high noise in our experimental setup. We find that, besides MCInfoNCE, ELK also recovers $\kappa(x)$ well (Rank Corr. $=0.92$). This is the first confirmation that ELK predicts correct posteriors in a controlled setup and opens space for future theoretical investigations.

\subsection{Posteriors\hspace{-0.15mm} Reflect\hspace{-0.15mm} Aleatoric\hspace{-0.15mm} Uncertainty\hspace{-0.15mm} in\hspace{-0.15mm} Practice}

After confirming that the predicted posteriors are correct, this section shows that they resemble the aleatoric uncertainty in image data. We also show that this enables novel applications such as credible intervals for image retrieval. 

\begin{table}[t]
    \centering
    \caption{Predicted certainties $\hat{\kappa}(x)$ of MCInfoNCE correlate with human annotator disagreement and information reduction via cropping images smaller. Rank correlation on unseen test data.}
    \label{tab:bench_cifar}
    \resizebox{0.33\textwidth}{!}{
    \begin{tabular}{lcc}
    \toprule
         Loss & Annotator Entropy $\uparrow$ & Crop Size $\uparrow$ \\
         \midrule
         HIB & \res{0.28}{0.00} & \res{0.69}{0.02} \\
         ELK & \res{0.14}{0.05} & \res{0.51}{0.03}\\
         MCInfoNCE & \res{0.29}{0.01} & \res{0.68}{0.01}\\
         \bottomrule
    \end{tabular}
    }
\end{table}

\begin{figure}[t]
    \centering
    \includegraphics{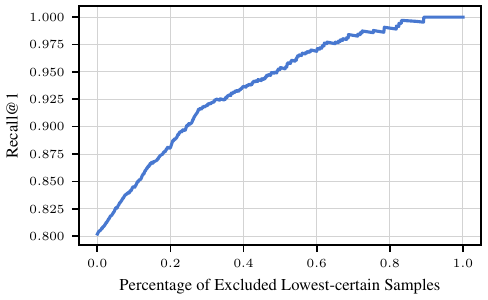} 
    \caption{Rejecting images with low certainty values $\hat{\kappa}(x)$ improves the performance on the remaining data monotonically with the threshold. This shows that $\hat{\kappa}(x)$ is predictive of performance.}
    \label{fig:erc}
\end{figure}

\textbf{Measuring Aleatoric Uncertainty.} In the upcoming experiment, we do not have access to any ground-truth $\kappa(x)$ against which to compare $\hat{\kappa}(x)$. Instead, we need to compare it to various indicators of aleatoric uncertainty. We use three different indicators that capture human uncertainty, information loss, and performance decrease with respect to the amount of aleatoric uncertainty. First, if an image is ambiguous, human annotators disagree about the latent that it shows. We therefore conduct our experiment on CIFAR-10H \cite{peterson2019human}. It comprises fifty class annotations for each image. This gives a soft-label distribution whose entropy reflects the ambiguity of the image. We compute the Rank Corr. between $1/\hat{\kappa}(x)$ and this annotator entropy to measure how well $\hat{\kappa}(x)$ reflects human-perceived input ambiguity. Second, we induce controlled information loss by deteriorating the image. \citep{wu2020simple} identified cropping to increase aleatoric uncertainty most clearly. Thus, we crop test images to percentages $\texttt{crop\_size} \sim \text{Unif}([0.25, 1])$ of their original size. The aleatoric uncertainty increases the more the image is cropped. We thus report the Rank Corr. between $1/\hat{\kappa}(x)$ and the crop size as a second metric. Third, ambiguous images inevitably lead to decreased performance. To investigate whether $\hat{\kappa}(x)$ is indicative of performance, we calculate the Recall@1 \cite{jegou2010product} on the $p\%$ images with the highest $\hat{\kappa}(x)$. If $\hat{\kappa}(x)$ correctly reflects aleatoric uncertainty, removing ambiguous images should improve performance, so the Recall@1 should increase monotonically with $p$. This metric also illustrates the popular use case of abstaining from uncertain predictions.

\textbf{Architecture and Training.} We translate the CIFAR-10H classification task into a contrastive task by considering images to be positive if they are in the same class and negative otherwise. We create training examples $(x, x^+, x^-_1, \dotsc, x^-_M)$ by drawing class labels for each image from its soft class distribution, selecting a random image $x$, an image $x^+$ with the same class label, and $M$ images $x^-_m$ with different class labels. On this data, we train a ResNet-18 \cite{he2016deep} pre-trained on CIFAR-10 \cite{huy_phan_2021_4431043} that outputs embeddings $e(x)$. We define $\hat{\mu}(x) := e(x)/\lVert e(x)\rVert_2$ and, following common practices for probabilistic embeddings \cite{kirchhof2022non,scott2021mises,li2021spherical}, $\hat{\kappa}(x)$ as $\lVert e(x)\rVert_2$. We run a 5-fold cross validation where we train for 175 epochs and select the best epoch via the Rank Corr. with the crop size on validation data. We choose this metric over the others because it can be computed on any dataset without additional supervision. All details on generating the contrastive data and the hyperparameter search are in the supplementary.

\begin{table}[t]
    \centering
    \caption{$\hat{\kappa}(x)$ can be learned by MCInfoNCE from both soft and hard labels. Rank correlation on unseen test data.}
    \label{tab:cifar_dataset}
    \resizebox{0.41\textwidth}{!}{
    \begin{tabular}{lcc}
    \toprule
         Labels & Annotator Entropy $\uparrow$ & Crop Size $\uparrow$ \\
         \midrule
         CIFAR-10H Soft Labels & \res{0.29}{0.01} & \res{0.68}{0.01} \\
         CIFAR-10H Hard Labels & \res{0.24}{0.01} & \res{0.64}{0.02} \\
         CIFAR-10 Hard Labels & \res{0.28}{0.01} & \res{0.69}{0.02} \\
         \bottomrule
    \end{tabular}
    }
\end{table}

\textbf{Results.} \cref{tab:bench_cifar} shows that $\hat{\kappa}(x)$ learned via MCInfoNCE has a high Rank Corr. of $0.68$ with the information lost due to cropping, i.e., images with less information return more uncertain posteriors. The correlation with the human annotator entropy is lower (0.29), but positive. HIB achieves a similar performance, while ELK shows lower correlations with both ground-truths (0.51 and 0.14, resp.). \cref{fig:erc} shows the performance decrease metric. Up to noise, the Recall@1 increases monotonically as images with the lowest $\hat{\kappa}(x)$ are rejected. This means that $\hat{\kappa}(x)$ is a good predictor of performance. As an additional qualitative metric the supplementary shows images with the lowest and highest $\hat{\kappa}(x)$ of each class. MCInfoNCE learns from labeling noise in this experiment, since the image class was drawn anew from its soft label distribution each time the image was used. In practice, we may have only one annotation per image, so that labeling noise occurs across examples rather than on each individual image. To this end, we further train on hard labels. These are either the most likely class of each soft label distribution on CIFAR-10H or the classical class labels on the CIFAR-10. \cref{tab:cifar_dataset} shows that MCInfoNCE can learn under both of these circumstances with a performance roughly equal to that when soft labels are available.

\begin{figure}[]
    \centering
    \begin{tikzpicture}
    \node[anchor=south west,inner sep=0] (image) at (0,0) {\includegraphics[trim=0cm 0.4cm 0cm -0.35cm, width=0.47\textwidth,page=2]{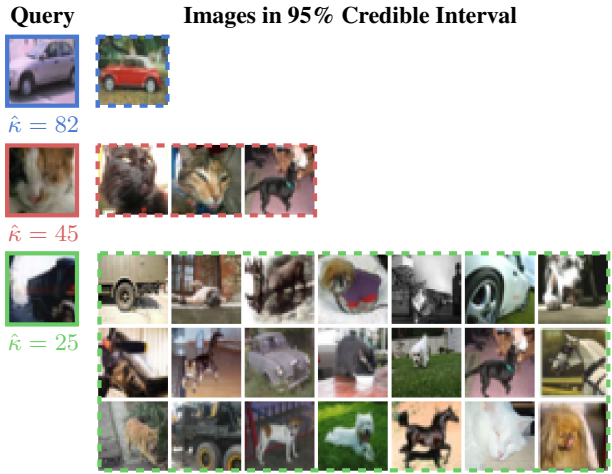}};
    \begin{scope}[x={(image.south east)},y={(image.north west)}]
        \node [anchor=south west] at (-0.005,0.95) {\footnotesize \textbf{Query}};
        \node [anchor=south] at (0.57,0.95) {\footnotesize \textbf{Images in 95\% Credible Interval}};
        \draw[blue,ultra thick] (0.005,0.951) rectangle (0.12,0.795);
        \draw[blue,ultra thick,dashed] (0.154,0.951) rectangle (0.269,0.795);
        \node [anchor=north west] at (-0.009,0.795) {\footnotesize \textcolor{blue}{$\hat{\kappa} = 82$}};
        \draw[red,ultra thick] (0.005,0.705) rectangle (0.12,0.547);
        \draw[red,ultra thick,dashed] (0.154,0.705) rectangle (0.512,0.547);
        \node [anchor=north west] at (-0.009,0.547) {\footnotesize \textcolor{red}{$\hat{\kappa} = 45$}};
        \draw[green,ultra thick] (0.005,0.461) rectangle (0.12,0.3);
        \draw[green,ultra thick,dashed] (0.157,0.461) rectangle (0.995,-0.03);
        \node [anchor=north west] at (-0.009,0.3) {\footnotesize \textcolor{green}{$\hat{\kappa} = 25$}};
    \end{scope}
    \end{tikzpicture}
    \caption{We use an image's posterior to define the credible interval that its latents lie in with a given probability. Clear query images (top) have small credible intervals containing images of the same class as the query. More ambiguous queries (bottom) return larger credible intervals with images from multiple possible classes.}
    \label{fig:unc_retrieval}
\end{figure}

\textbf{Credible Intervals for Image Retrieval.} Since we estimate posteriors $Q(z|x)$, we can also introduce Bayesian credible intervals \cite{lee1989bayesian} to our image representation task. Such intervals $\text{CI}_p(x) \subset \mathcal{Z}$ contain the true generative latent $z$ of $x$ with a user-defined probability $p \in [0, 1]$, i.e., $P(z \in \text{CI}_p(x)) = p$ for $x \sim P(x|z)$. Credible intervals help understand the degree to which our model can identify the latent that $x$ shows. We can visualize these latents by searching for images whose $\hat{\mu}(x)$ fall within $\text{CI}_p$. \cref{fig:unc_retrieval} shows such intervals on our MCInfoNCE model for CIFAR-10H. A clear image (top) has a sharp posterior and thus a small CI containing only one image from the same class. The CI of a more ambiguous query image, like the second, tells us that the model places the query in the region of cats, but that it could also be a dog. Highly ambiguous queries, like the last one, lead to wide CIs that span multiple possible classes. They examples show how credible intervals can augment retrieval with uncertainty-awareness: They determine the number of images to retrieve subject to the query's ambiguity and allow users to judge the uncertainty better than a simple scalar uncertainty value. 

\section{Discussion}

\textbf{Relations to Broader Variational Inference.} Our work advances the recent theoretical discussions about contrastive learning and variational inference. \citet{oord2018representation} and \citet{,poole2019variational} initially showed that the minimizer of InfoNCE is the likelihood ratio of positive and negative densities of the generative process. \citet{dml_inversion} used this to show that the minimizer recovers the latents, modulo rotations. Our work shows that we can even learn the correct posterior of a \textit{probabilistic} generative process, modulo rotations, i.e., the internal probabilistic latent representations of our specific encoder are indeed \textit{correct}. This may have implications to other works on variational approaches and contrastive learning, like \citet{aitchison2021infonce}.

\textbf{Multi-modal Posteriors.} The vMF posteriors should be able to capture most augmentations in self-supervised contrastive learning that deteriorate the image whole image, i.e., all latent factors equally. However, it is also interesting to think about deteriorations that lead to multi-modal posteriors. In this case, \cref{form:equal_marginals} does not make any parametric assumption on the posteriors and thus still holds. \cref{eq:mono} and \cref{eq:equalkappa} need to be extended regarding the identifiability of the mixture component, but could then utilize our propositions for each component. We see this as an exciting direction for future works.

\section{Conclusion}

This work presented MCInfoNCE, a probabilistic contrastive loss that predicts posteriors instead of points. We proved that it learns the generative processes' true posteriors. This provides a theoretical grounding for the recent probabilistic embeddings literature and connects it to a probabilistic extension of nonlinear ICA. In practice, the posteriors allow predicting the level of aleatoric uncertainty in ambiguous inputs as well as estimating credible intervals with flexible sizes depending on a query's ambiguity in image retrieval. These are only two usages that correct posteriors enable and further usages are a promising area for future research. Aleatoric uncertainty is not only faced in computer vision and retrieval. We hope that the blueprint way of enhancing InfoNCE into MCInfoNCE inspires applications in further tasks with intrinsic ambiguities in their inputs. 

\section*{Acknowledgements} 
Kay Choi has helped designing Figure 1. This work was funded by the Deutsche Forschungsgemeinschaft (DFG, German Research Foundation) under Germany's Excellence Strategy -- EXC number 2064/1 -- Project number 390727645. The authors thank the International Max Planck Research School for Intelligent Systems (IMPRS-IS) for supporting Michael Kirchhof.

\bibliography{icml2022}
\bibliographystyle{icml2023}

\newpage
\appendix
\onecolumn
\section{Proofs}

\subsection{Proof of \cref{form:equal_marginals}}

\textbf{\cref{form:equal_marginals}} ($\mathcal{L}$ is minimized iff marginals match)
    Let the latent marginal distributions $P(z) = \int P(z|x) dP(x)$ and $\int Q(z|x) dP(x)$ be uniform. $\lim_{M \rightarrow \infty}\mathcal{L}$ attains its minimum when $\forall x, x^+ \in \{x \in \mathcal{X} | P(x) > 0\}$
    \begin{align*}
       & \iint Q(z|x) Q(z^+|x^+) P(z^+|z) dz^+ dz = \\
       & \iint P(z|x) P(z^+|x^+) P(z^+|z) dz^+ dz \,\,.
    \end{align*}

\textbf{Proof.} All of the above densities are integrable, so we can write the loss function $\mathcal{L}$ in the form of Riemann integrals.
\begin{align}
    \lim\limits_{M \rightarrow \infty} \mathcal{L} = -\lim\limits_{M \rightarrow \infty} \int & P(x) P(x^+|x) \int \prod\limits_{m=1}^M P(x_m^-) \log \int Q(z | x) Q(z^+ | x^+) \\
    & \prod\limits_{m=1}^M Q(z^-_m|x^-_m) \frac{e^{\kappa_\text{pos} z^\top z^+}}{\frac{1}{M} e^{\kappa_\text{pos} z^\top z^+} + \frac{1}{M} \sum\limits_{m=1}^M e^{\kappa_\text{pos} z^\top z^-_m}} dz^-_1 \dotsc z^-_M dz^+ dz dx_1^- \dotsc dx_M^- dx^+ dx
\end{align}
We know that $\kappa_\text{pos} < \infty$, $\kappa(x) < \infty \,\forall x\in\mathcal{X}$, the normalization constants $C(\kappa) < \infty \,\forall \kappa < \infty$, and the dot products are bounded. This implies that all densities inside these integrals as well as the exponentials in the fraction are bounded. Thus, the whole term inside the outmost integral is bounded. Due to the dominated convergence theorem we can pull the limit into the integral.
\begin{align}
    = -\int & P(x) P(x^+|x) \lim\limits_{M \rightarrow \infty} \int \prod\limits_{m=1}^M P(x_m^-) \log \int Q(z | x) Q(z^+ | x^+) \\
    & \prod\limits_{m=1}^M Q(z^-_m|x^-_m) \frac{e^{\kappa_\text{pos} z^\top z^+}}{\frac{1}{M} e^{\kappa_\text{pos} z^\top z^+} + \frac{1}{M} \sum\limits_{m=1}^M e^{\kappa_\text{pos} z^\top z^-_m}} dz^-_1 \dotsc z^-_M dz^+ dz dx_1^- \dotsc dx_M^- dx^+ dx
\end{align}
The strong law of large numbers and the fact that $\int Q(z^- | x^-) P(x^-) dx^- = P(z)$ imply
\begin{align}
    = -\int & P(x) P(x^+|x) \lim\limits_{M \rightarrow \infty} \log \int Q(z | x) Q(z^+ | x^+) \frac{e^{\kappa_\text{pos} z^\top z^+}}{\frac{1}{M} e^{\kappa_\text{pos} z^\top z^+} + \mathop{\mathbb{E}}\limits_{z^- \sim P(z)}(e^{\kappa_\text{pos} z^\top z^-})} dz^+ dz dx^+ dx\,\,.
\end{align}
Both densities and the fraction inside the inner integral are positive and bounded, so the integral is, too. In this range, i.e., $(0, \infty)$, the logarithm is continuous, so the continuous mapping theorem gives
\begin{align}
    = -\int & P(x) P(x^+|x) \log \lim\limits_{M \rightarrow \infty} \int Q(z | x) Q(z^+ | x^+) \frac{e^{\kappa_\text{pos} z^\top z^+}}{\frac{1}{M} e^{\kappa_\text{pos} z^\top z^+} + \mathop{\mathbb{E}}\limits_{z^- \sim P(z)}(e^{\kappa_\text{pos} z^\top z^-})} dz^+ dz dx^+ dx\,\,.
\end{align}
With the arguments from above, the inside of the inner integral is bounded, so we can again apply the dominated convergence theorem.
\begin{align}
    & = -\int P(x) P(x^+|x) \log \int Q(z | x) Q(z^+ | x^+) \lim\limits_{M \rightarrow \infty} \frac{e^{\kappa_\text{pos} z^\top z^+}}{\frac{1}{M} e^{\kappa_\text{pos} z^\top z^+} + \mathop{\mathbb{E}}\limits_{z^- \sim P(z)}(e^{\kappa_\text{pos} z^\top z^-})} dz^+ dz dx^+ dx \\
    & = -\int P(x) P(x^+|x) \log \int Q(z | x) Q(z^+ | x^+) \frac{e^{\kappa_\text{pos} z^\top z^+}}{\mathop{\mathbb{E}}\limits_{z^- \sim P(z)}(e^{\kappa_\text{pos} z^\top z^-})} dz^+ dz dx^+ dx
\end{align}
Since $P(z) = \text{Unif}(\mathcal{S}^{D - 1}) = \frac{1}{\lVert \mathcal{S}^{D-1}\rVert}$, which we define as $\frac{1}{S}$ in shorthand, we get
\begin{align}
    & = -\int P(x) P(x^+|x) \log S \int Q(z | x) Q(z^+ | x^+) \frac{e^{\kappa_\text{pos} z^\top z^+}}{\int\limits_{\mathcal{S}^{D-1}} e^{\kappa_\text{pos} z^\top z^-} dz^-} dz^+ dz dx^+ dx  \\ \label{form:proof_a}
    & = -\int P(x) P(x^+|x) \log S \int Q(z | x) Q(z^+ | x^+) P(z^+|z) dz^+ dz dx^+ dx  \,\,.
\end{align}
Let us turn our attention to $P(x^+|x)$. By marginalization, factorization, and the conditional independencies of the data-generating process, we get
\begin{align}
    & P(x^+|x) \\
    =& \int P(x^+, z^+, z | x) dz^+ dz\\
    =& \int P(x^+|z^+, z, x) P(z^+|z, x) P(z|x) dz^+ dz \\
    =& \int P(x^+ | z^+) P(z^+ | z) P(z|x) dz^+ dz \,\,.
\end{align}
After a multiplication with 1, Bayes Theorem, and using $P(z) = \frac{1}{S}$, we get 
\begin{align}
    =& \int \frac{P(x^+ | z^+) P(z^+) P(x^+)}{P(z^+) P(x^+)} P(z^+ | z) P(z|x) dz^+ dz \\
    =& \int P(z|x) P(z^+ | x^+) P(z^+ | z) \frac{P(x^+)}{P(z^+)} dz^+ dz \\
    =& P(x^+)\, S \int P(z|x) P(z^+ | x^+) P(z^+ | z)dz^+ dz \,\,.
\end{align}
We can insert this into Formula \ref{form:proof_a}.
\begin{align}
    & -\int P(x) P(x^+)\, S \int P(z|x) P(z^+ | x^+) P(z^+ | z)dz^+ dz \\
    & \log S \int Q(z | x) Q(z^+ | x^+) P(z^+|z) dz^+ dz dx^+ dx \\
    = & \mathop{\mathbb{E}}\limits_{\substack{x \sim P(x) \\ x^+ \sim P(x^+)}}\left( S \int P(z|x) P(z^+ | x^+) P(z^+ | z)dz^+ dz \log S \int Q(z | x) Q(z^+ | x^+) P(z^+|z) dz^+ dz \right)\,.
\end{align}
Note that both terms are conditional on $x, x^+$ and the expected value is taken over both of these. I.e., $\mathcal{L}$ in the limit is a (non-normalized) cross-entropy between $\int P(z|x) P(z^+ | x^+) P(z^+ | z)dz^+ dz$ and $\int Q(z | x) Q(z^+ | x^+) P(z^+|z) dz^+ dz$. The loss is minimized iff the two terms match for all values in the outmost expected value, i.e., $\forall x, x^+ \in \{x \in \mathcal{X} | P(x) > 0\}$. $\hfill\square$

\subsection{Proof of \cref{eq:mono}}

\textbf{\cref{eq:mono}} (The marginal is a function)
    Let $P(z|x)$ and $P(z^+|z)$ be vMF distributions as defined in \cref{sec:setup}. Given $x, x^+ \in \mathcal{X}$, we can rewrite \begin{align}
        & \iint P(z|x) P(z^+|x^+) P(z^+|z) dz^+ dz \\
        & =: h_{\kappa_\text{pos}}(\mu(x)^\top \mu(x^+), \kappa(x), \kappa(x^+)), 
    \end{align} 
    i.e., as a function $h_{\kappa_\text{pos}}$ that depends only on $\mu(x)^\top \mu(x^+), \kappa(x),$ and $\kappa(x^+)$. The same function can be used for $\hat{\mu}(x)^\top \hat{\mu}(x^+), \hat{\kappa}(x), \hat{\kappa}(x^+)$:
    \begin{align}
        & \iint Q(z|x) Q(z^+|x^+) P(z^+|z) dz^+ dz \\
        & = h_{\kappa_\text{pos}}(\hat{\mu}(x)^\top \hat{\mu}(x^+), \hat{\kappa}(x), \hat{\kappa}(x^+)). 
    \end{align}

\textbf{Proof.} Let us first insert the vMF densities.
\begin{align}
     & \iint P(z|x) P(z^+|x^+) P(z^+|z) dz^+ dz \\
    =& C(\kappa(x^+)) C(\kappa_\text{pos}) \iint C(\kappa(x)) \exp[\kappa(x) \mu(x)^\top z + \kappa(x^+) \mu(x^+)^\top z^+ + \kappa_\text{pos} z^\top z^+] dz^+ dz \\
    =& C(\kappa(x^+)) C(\kappa_\text{pos}) \int C(\kappa(x)) \exp(\kappa(x) \mu(x)^\top z) \int \exp[(\kappa(x^+) \mu(x^+) + \kappa_\text{pos} z)^\top z^+] dz^+ dz
\end{align}
The term inside the inner integral can be rewritten into an unnormalized vMF density if we specify $\mu^* := \frac{\kappa(x^+) \mu(x^+) + \kappa_\text{pos} z}{\lVert \kappa(x^+) \mu(x^+) + \kappa_\text{pos} z \rVert}$ and $\kappa^* := \lVert \kappa(x^+) \mu(x^+) + \kappa_\text{pos} z \rVert$. The integral over this density is 1.
\begin{align}
    =& C(\kappa(x^+)) C(\kappa_\text{pos}) \int C(\kappa(x)) \exp(\kappa(x) \mu(x)^\top z) \frac{1}{C(\kappa^*)} \int C(\kappa^*) \exp[\kappa^* {\mu^*}^\top z^+] dz^+ dz \\
    =& C(\kappa(x^+)) C(\kappa_\text{pos}) \int C(\kappa(x)) \exp(\kappa(x) \mu(x)^\top z) \frac{1}{C(\kappa^*)} dz \\
    =& C(\kappa_\text{pos}) \mathop{\mathbb{E}}\limits_{z \sim \text{vMF}(\mu(x), \kappa(x))} \left(\frac{C(\kappa(x^+))}{C\left(\sqrt{\kappa(x^+)^2 + \kappa_\text{pos}^2 + 2\kappa(x^+) \kappa_\text{pos} \mu(x^+)^\top z}\right)}\right) \label{form:h} \\
    =: & h_{\kappa_\text{pos}}(\mu(x)^\top \mu(x^+), \kappa(x), \kappa(x^+))
\end{align}
In the last step, the expected value is over $\mu(x^+)^\top z, z \sim \text{vMF}(\mu(x), \kappa(x))$. This depends only on the distance $\mu(x)^\top \mu(x^+)$ instead of the full location parameters $\mu(x)$ and $\mu(x^+)$ because the vMF is rotationally symmetric and we can perform a suitable Householder rotation, see also \citet{romanazzi2014discriminant}. $\hfill \square$

\subsection{Proof of \cref{eq:equalkappa}}

\textbf{\cref{eq:equalkappa}} (Arguments of $h_\text{pos}$ must be equal)
    Define $h_\text{pos}$ as in \cref{eq:mono}. Let $\mathcal{X}' \subseteq \mathcal{X}$, $\mu, \hat{\mu}: \mathcal{X}' \rightarrow \mathcal{Z}$, $\kappa, \hat{\kappa}: \mathcal{X}' \rightarrow \mathbb{R}_{> 0}$, $\kappa_\text{pos} > 0$. If $h_\text{pos}(\hat{\mu}(x)^\top \hat{\mu}(x^+), \hat{\kappa}(x), \hat{\kappa}(x^+)) = h_\text{pos}(\mu(x)^\top \mu(x^+), \kappa(x), \kappa(x^+))$ $\forall x, x^+ \in \mathcal{X}'$, then 
    \begin{align}
        & \hat{\mu}(x)^\top \hat{\mu}(x^+) = \mu(x)^\top \mu(x^+) \text{ and} \\
        & \hat{\kappa}(x) = \kappa(x) \,\,\, \forall x, x^+ \in \mathcal{X}'.
    \end{align}

\textbf{Proof.} (a) The normalization constant of the vMF $C(\kappa) = \frac{\kappa^{D/2 - 1}}{(2\pi)^{D/2} I_{D/2 -1}(\kappa)}$, where $I_o$ is the modified Bessel function of the first kind and order $o$, is strictly monotonically decreasing and convex \cite{kirchhof2022non}. 

(b) Consider arbitrary $x = x^+$, $x \in \mathcal{X}'$. In this case, $\mu(x)^\top \mu(x^+) = \hat{\mu}(x)^\top \hat{\mu}(x^+) = 1$, and both sides of the equality simplify
\begin{align}
     \iint Q(z|x) Q(z^+|x^+) P(z^+|z) dz^+ dz &=
    \iint P(z|x) P(z^+|x^+) P(z^+|z) dz^+ dz \\
    \iff  h_{\kappa_\text{pos}}(1, \kappa(x), \kappa(x)) &= h_{\kappa_\text{pos}}(1, \hat{\kappa}(x), \hat{\kappa}(x)) \\
    \iff  \tilde{h}_{\kappa_\text{pos}}(\kappa(x)) &= \tilde{h}_{\kappa_\text{pos}}(\hat{\kappa}(x)) \label{form:equality}
\end{align}
with $\tilde{h}_{\kappa_\text{pos}}(\kappa) := h_{\kappa_\text{pos}}(1, \kappa, \kappa)$. Due to (a), the denominator in Formula~\ref{form:h} grows strictly faster than the numerator. So $\tilde{h}$ is strictly monotonically increasing. Thus, $\tilde{h}_{\kappa_\text{pos}}(\kappa(x)) = \tilde{h}_{\kappa_\text{pos}}(\hat{\kappa}(x))$ only if $\kappa(x) = \hat{\kappa}(x)$.

(c) Let $x, x^+ \in \mathcal{X}'$ be arbitrary. From (b) we know $\hat{\kappa}(x) = \kappa(x)$, so we can simplify
\begin{align}
    h_{\kappa_\text{pos}}(\mu(x)^\top \mu(x^+), \kappa(x), \kappa(x^+)) &= h_{\kappa_\text{pos}}(\hat{\mu}(x)^\top \hat{\mu}(x^+), \hat{\kappa}(x), \hat{\kappa}(x^+)) \\
    \iff  h^*_{\kappa_\text{pos}, \kappa(x), \kappa(x^+)}(\mu(x)^\top \mu(x^+)) &= h^*_{\kappa_\text{pos}, \kappa(x), \kappa(x^+)}(\hat{\mu}(x)^\top \hat{\mu}(x^+)) 
\end{align}
with $h^*_{\kappa_\text{pos}, \kappa(x), \kappa(x^+)}(\cdot) := h_{\kappa_\text{pos}}(\cdot, \kappa(x), \kappa(x^+))$. In other words, both sides of the equality are the same function $h^*_{\kappa_\text{pos}, \kappa(x), \kappa(x^+)}$ with only one free variable. Due to (a), the denominator in Formula \ref{form:h} strictly decreases with increasing $\mu(x)^\top \mu(x^+)$ if $\kappa(x^+) > 0$ and $\kappa_\text{pos} > 0$. So, $h^*_{\kappa_\text{pos}, \kappa(x), \kappa(x^+)}$ is strictly monotonically increasing and $h^*_{\kappa_\text{pos}, \kappa(x), \kappa(x^+)}(\mu(x)^\top \mu(x^+)) = h^*_{\kappa_\text{pos}, \kappa(x), \kappa(x^+)}(\hat{\mu}(x)^\top \hat{\mu}(x^+))$ implies $\mu(x)^\top \mu(x^+) = \hat{\mu}(x)^\top \hat{\mu}(x^+)$. $\hfill \square$

\subsection{Proof of \cref{eq:final}}

\textbf{\cref{eq:final}} ($\mathcal{L}$ identifies the correct posteriors)
Let $\mathcal{Z} = \mathcal{S}^{D-1}$ and $P(z) = \int P(z|x) dP(x)$ and $\int Q(z|x) dP(x)$ be the uniform distribution over $\mathcal{Z}$. Let $g$ be a probabilistic generative process defined in Formulas \ref{form:gen1}, \ref{form:gen2}, and \ref{form:gen3} with known $\kappa_\text{pos}$. Let $g$ have vMF posteriors $P(z|x) = \text{vMF}(z;\mu(x), \kappa(x))$ with $\mu: \mathcal{X} \rightarrow \mathcal{S}^{D-1}$ and $\kappa: \mathcal{X} \rightarrow \mathbb{R}_{> 0}$. Let an encoder $f(x)$ parametrize vMF distributions $\text{vMF}(z; \hat{\mu}(x), \hat{\kappa}(x))$. Then $f^* = \arg\min_f \lim_{M \rightarrow \infty} \mathcal{L}$ has the correct posteriors up to a rotation of $\mathcal{Z}$, i.e., $\hat{\mu}(x) = R \mu(x)$ and $\hat{\kappa}(x) = \kappa(x)$, where $R$ is an orthogonal rotation matrix, $\forall x \in \{x \in \mathcal{X} | P(x) > 0\}$.

\textbf{Proof.} If $f^*$ optimizes $\mathcal{L}$, then by \cref{form:equal_marginals} $\forall x, x^+ \in \{x \in \mathcal{X} | P(x) > 0\}$ we have 
\begin{align}
       \iint Q(z|x) Q(z^+|x^+) P(z^+|z) dz^+ dz = \iint P(z|x) P(z^+|x^+) P(z^+|z) dz^+ dz \,\,.
\end{align}
Then by \cref{eq:equalkappa} with $\mathcal{X}' := \{x \in \mathcal{X} | P(x) > 0\}$ we get $\hat{\kappa}(x) = \kappa(x)$ and $\mu(x)^\top \mu(x^+) = \hat{\mu}(x)^\top \hat{\mu}(x^+)$. With the extended Mazur-Ulam Theorem \cite{dml_inversion}, the latter implies $\hat{\mu}(x) = R \mu(x)$ with an orthogonal rotation matrix $R \in \mathbb{R}^{D \times D}$. $\hfill \square$

\section{Controlled Experiment}

\subsection{Network Architectures}

We use MLPs to parametrize the generative processes' posteriors $\mu(x)$ and $\kappa(x)$ as well as the encoder $\hat{\mu}(x)$ and $\hat{\kappa}(x)$. 

For $\mu(x)$ and $\hat{\mu}(x)$ we follow \citet{dml_inversion}. The MLP for $\mu(x)$ has three linear layers with 10 dimensions and leaky ReLU activations. To prevent collapsed initializations we take $1000$ exemplary samples for $\mu(x)$ and re-initiate it if the smallest cosine similarity $x_1^\top x_2$ between any pair $x_1, x_2$ of them is bigger than $0.5$. $\hat{\mu}(x)$ has six hidden linear layers with leaky ReLU activations plus an input and and output layer with the input and output dimensions $[D \rightarrow 10 \cdot D, 10 \cdot D \rightarrow 50 \cdot D, 50 \cdot D \rightarrow 50 \cdot D, 50 \cdot D \rightarrow 50 \cdot D, 50 \cdot D \rightarrow 50 \cdot D, 50 \cdot D \rightarrow 50 \cdot D, 50 \cdot D \rightarrow 10 \cdot D, 10 \cdot D \rightarrow D]$. The outputs of both networks are normalized to an $L_2$ norm of $1$ to ensure they are on the unit sphere. 

The MLPs for $\kappa(x)$ and $\hat{\kappa}(x)$ have the same architecture as $\mu(x)$ and $\hat{\mu}(x)$, but $\kappa(x)$ has one less hidden layer than $\mu(x)$. The last layer of both networks outputs only a scalar instead of a $D$-dimensional vector. It is postprocessed by $\tilde{\kappa}(x) = 1 + \exp(\kappa(x))$ to ensure their strict positivity. Before training, $\hat{\kappa}(x)$ is normalized to output the same range of values as $\kappa(x)$ to improve training stability.

\subsection{Generating Contrastive Training Data}

The generative process in \cref{sec:setup} first draws latents $z$ and then generates observations $x$ to create contrastive training data. However, we want to control our generative processes' posteriors. Thus, we need to first sample $x$ and then $z \sim P(z|x)$. A method to sample backwards like this while still obtaining samples as if they were from the forward generative process is rejection sampling. We first draw random candidates $(x, x^+)$ from $\mathcal{X} = [0, 1]^D$, then draw $(z, z^+)$ from their corresponding posteriors. To ensure that they form a valid positive example as per the distributions in Formulas~\ref{form:gen1} and \ref{form:gen2}, we accept or reject them with a probability proportional to 
\begin{align}
    \frac{C(\kappa_\text{pos}) e^{\kappa_\text{pos} z^\top z^+}}{C(\kappa_\text{pos}) e^{\kappa_\text{pos} z^\top z^+} + C(0)}\,.
\end{align}
This is the probability that $z$ and $z^+$ are positive to one another. The proposal distribution's density for rejection sampling is dropped here due to the uniform priors. Negative examples $(x^-_m)_{m=1, \dotsc, M}$ are drawn randomly from $\mathcal{X}$ due to Formula~\ref{form:gen3}.

\subsection{Experiment Parameters}

Following \citet{dml_inversion}, all experiments used $\kappa_\text{pos} = 20$ and the above network architectures. The learning rate was $0.0001$ and was decreased after each $25\%$ of training progress by a factor of $0.1$. Performance was measured at the end of the training without early stopping on $10000$ sampled $x$ points. All experiments were implemented in Python 3.8.11, PyTorch 1.9.0 on NVIDIA-RTX 2080TI GPUs with 12GB VRAM. \cref{tab:contr_exp_params} below summarizes the remaining parameters used by all ablations of the controlled experiment.

\begin{table}[h]
    \centering
    \resizebox{\textwidth}{!}{
    \begin{tabular}{lccccccccl}
        \toprule
         Experiment & Gen. $D$ & Enc. $D'$ & Posterior & $\min(\kappa(x))$ & $\max(\kappa(x))$ & Batchsize & Number of Batches & Number MC Samples & Comment\\
         \midrule
         Ambiguous ($\kappa(x) \in [16, 32]$) & 10 & 10 & vMF & 16 & 32 & 512 & 100000 & 512 & Also used for HIB, ELK, InfoNCE\\
         Clear ($\kappa(x) \in [64, 128]$) & 10 & 10 & vMF & 64 & 128 & 512 & 100000 & 512 & \\
         Injective ($\kappa(x) = \infty$) & 10 & 10 & vMF/Dirac & $\infty$ & $\infty$ & 512 & 100000 & 512 & \\
         $D = 2$ & 2 & 2 & vMF & 16 & 32 & 512 & 8192 & 512 & \\
         Gaussian & 10 & 10 & Gaussian & 16 & 32 & 512 & 100000 & 512 & $\sigma^2 = 1/\kappa(x)$ \\
         Laplace & 10 & 10 & Laplace & 16 & 32 &  512 & 100000 & 512 & $b = 1/\kappa(x)$ \\
         MC Samples & 10 & 10 & vMF & 16 & 32 & 512 & 100000 & $x$ & $x \in \{ 1, 4, 16, 64, 256, 512\}$\\
         Encoder Dim & 10 & $x$ & vMF & 16 & 32 & 512 & 100000 & 512 & for $x \in \{ 4, 8, 10, 16, 32\}$\\
         --- \raisebox{-0.5ex}{''} --- & & & & & & 512 & & 256 & for $x = 64$ \\
         --- \raisebox{-0.5ex}{''} ---  & & & & & & 256 & & 256 & for $x = 128$ \\
         High Dim & $x$ & $x$ & vMF & 16 & 32 & 512 & 100000 & 512 & $x \in \{ 10, 16 \}$\\
         --- \raisebox{-0.5ex}{''} --- & & & & & & 256 & & 256 & for $x \in \{ 32, 40, 48, 56, 64 \}$ \\
         \bottomrule
    \end{tabular}
    }
    \caption{Parameters of the generative process and loss in the controlled experiments. $x$ denotes variable parameters. Batchsize and number of MC samples were reduced in high dimensions to not exceed the available VRAM.}
    \label{tab:contr_exp_params}
\end{table}

\subsection{Contrastive Hedged Instance Embeddings}

HIB \cite{oh2018modeling} is formulated similarly to MCInfoNCE in that it also draws samples of a posterior and computes a probability score with them. HIB originally uses Gaussians and compares $L_2$ distances between samples. We adapt this to vMFs and cosine distances to align it with the spherical formulation of the latent space. The reformulated HIB loss is 
\begin{align}
    \mathcal{L}_\text{HIB} := \hspace{-0.5cm} \mathop{\mathbb{E}} \limits_{\substack{x \sim P(x) \\ x^+ \sim P(x^+|x) \\ x^-_m \sim P(x^-), m=1, \dotsc, M}} \hspace{-0.5cm} \left( -\log \hspace{-0.5cm} \mathop{\mathbb{E}}\limits_{\substack{z \sim Q(z|x) \\ z^+ \sim Q(z^+|x^+)}} \hspace{-0.5cm} \left( s(a \cdot z^\top z^+ + b) \right) - \frac{1}{M} \sum\limits_{m=1}^{M} \log \hspace{-0.5cm}\mathop{\mathbb{E}}\limits_{\substack{z \sim Q(z|x) \\ z^+ \sim Q(z^-|x^-_m)}}\hspace{-0.5cm} \left(1 - s(a \cdot z^\top z^-_m + b) \right) \right) \hspace{-0.5mm},
\end{align}
where $s(\cdot)$ is the Sigmoid function and $a$ and $b$ are tuneable hyperparameters. We excluded the KL regularizer originally proposed by Oh et al. since none of the other losses receive prior information on $\kappa(x)$.

\subsection{Contrastive Expected Likelihood Kernel}

The ELK is commonly used inside a classification cross-entropy loss \cite{kirchhof2022non}. Its key characteristic is that it replaces the point-to-point distance, e.g., cosine distance, by the expected likelihood distance. An analytical solution to compare two vMFs is provided in the supplementary of Kirchhof et al.. We can plug this distance $d_\text{EL-vMF}(\hat{\mu}(x_1), \hat{\kappa}(x_1), \hat{\mu}(x_2), \hat{\kappa}(x_2))$ into InfoNCE and transform it into a similarity by multiplying it with $-1$ to obtain our contrastive ELK loss:
\begin{align}
    \mathcal{L}_\text{ELK} := \hspace{-1.2cm} \mathop{\mathbb{E}} \limits_{\substack{x \sim P(x) \\ x^+ \sim P(x^+|x) \\ x^-_m \sim P(x^-), m=1, \dotsc, M}} \hspace{-0.2cm} \left( -\log \frac{e^{-\kappa_\text{pos} d_\text{EL-vMF}(\hat{\mu}(x), \hat{\kappa}(x), \hat{\mu}(x^+), \hat{\kappa}(x^+))}}{\frac{1}{M} e^{- \kappa_\text{pos} d_\text{EL-vMF}(\hat{\mu}(x), \hat{\kappa}(x), \hat{\mu}(x^+), \hat{\kappa}(x^+))} + \frac{1}{M} \sum\limits_{m=1}^M e^{-\kappa_\text{pos} d_\text{EL-vMF}(\hat{\mu}(x), \hat{\kappa}(x), \hat{\mu}(x^-_m), \hat{\kappa}(x^-_m))}} \right) \hspace{-0.5mm}.
\end{align}

\subsection{Hyperparameter Tuning} \label{sec:hyp}

All losses were tuned on the "Standard" experiment setup via grid search. The seed for the generative process was exclusive and not used in the five seeds of the final results. \cref{tab:hyperparams} below gives the hyperparameters along with the chosen best setup according to the rank correlation between $\kappa(x)$ and $\hat{\kappa}(x)$.

\begin{table}[h]
    \centering
    \begin{tabular}{lccc}
    \toprule
         & HIB & ELK & MCInfoNCE \\
         \midrule
         Number of negatives $M$ & $\{\mathbf{0}, 1, 32\}$ & $\{0, \mathbf{1}, 32\}$& $\{0, 1, \mathbf{32}\}$ \\
         $\kappa_\text{pos}$ learnable & \{yes, \textbf{no}\} & \{yes, \textbf{no}\}&  \{yes, \textbf{no}\}\\
         Phasewise training & \{yes, \textbf{no}\}& \{yes, \textbf{no}\}& \{\textbf{yes}, no\}\\
         $a$ & $\{ 0.5, \mathbf{1}, 2, 4 \}$ \\
         $b$ & $\{-8, -4, -2, -1, \mathbf{0}, 1, 2, 4, 8\}$ \\
         \bottomrule
    \end{tabular}
    \caption{Possible hyperparameters and best-performing hyperparameters (\textbf{bold}). $M = 0$ corresponds to not sampling negatives, but using one sample from the same batch as a negative. HIB's additional hyperparameters were tuned after the first three parameters to reduce the number of grid-search evaluations.}
    \label{tab:hyperparams}
\end{table}

There are two interesting results in this tuning. First, the true generative $\kappa_\text{pos}$ was indeed the best choice. All methods performed worse when they learned it themselves (starting from the true value) or when given a different value (not shown here). Second, MCInfoNCE performs best with a high number of negative samples. This corroborates the theoretical study of its limiting behaviour as $M \rightarrow \infty$.

Phasewise training is the empirical strategy of first learning $\hat{\mu}(x)$ during the first half of epochs, then fixing it and learning $\hat{\kappa}(x)$ \cite{shi2019probabilistic,li2021spherical}. MCInfoNCE showed an improved performance with this strategy. This is likely because the training signal of $\kappa(x)$ is far lower in the loss than that of $\mu(x)$. During the training phase of $\hat{\mu}(x)$, it turned out beneficial to use negatives from the same batch, i.e., $M=0$. 

\subsection{Ablation with High Latent Space Dimension}

\begin{figure}[h]
    \centering
    \includegraphics{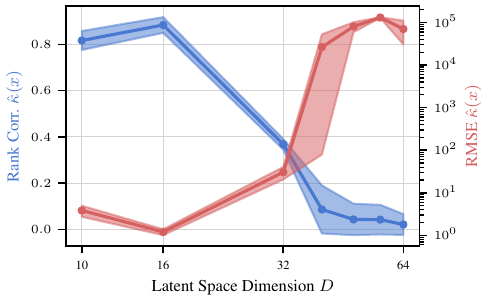}
    \caption{The metrics worsen if the generative process has a latent space of dimension $D \geq 40$. This is likely not due to MCInfoNCE, but a limitation of the contrastive setup of our controlled experiment. Mean $\pm$ std. err. for five seeds.}
    \label{fig:highdim}
\end{figure}

We use the latent space dimension $D=10$ for most experiments following \citet{dml_inversion}. Below in \cref{fig:highdim}, we increase the latent space dimension of the generative process and encoder up to $64$. We notice considerable performance drops for $D \geq 40$. Other losses than MCInfoNCE also suffer this. Hence, it is likely because of our experimental setup: We use uniformly distributed negatives instead of sophisticated negative mining and the rejection sampling has lower success probabilities in high dimensions, making it harder to generate valid contrastive examples. 

\subsection{Ablation with Joint Architecture} \label{sec:vae}

In the upper experiments, the networks for $\kappa(x)$ and $\mu(x)$ (and $\hat{\kappa}(x)$ and $\hat{\mu}(x)$) were independent, i.e., did not share parameters. This was to make clear that $\kappa(x)$ characterizes the uncertainty of the input $x$, rather than the latent of a shared backbone. However, a shared backbone with two heads for $\mu(x)$ and $\kappa(x)$ is a common architecture as, e.g., in VAEs. We've thus run an ablation where $\mu(x)$ is the output of the embedder (a 6-layer MLP) and $\kappa(x)$ is a 3-layer MLP attached after it. This keeps the total number of parameters the same as in the independent case. We rerun the "Ambiguous" setting with $\kappa(x) \in [16, 32]$. \cref{tab:ablation_shared} shows that MCInfoNCE achieves similar performance in both cases. 

\begin{table}[h]
    \centering
    \caption{MCInfoNCE also discovers correct posteriors if $\hat{\mu}(x)$ and $\hat{\kappa}(x)$ have a shared backbone. Mean $\pm$ std. err. for five seeds. \vspace{1mm}}
    \label{tab:ablation_shared}    
    \begin{tabular}{lcccc}
    \toprule
          &  \multicolumn{2}{c}{True vs Pred. Location $\hat{\mu}(x)$} & \multicolumn{2}{c}{True vs Pred. Certainty $\hat{\kappa}(x)$}\\
         Architecture & RMSE $\downarrow$ & Rank Corr. $\uparrow$ & RMSE $\downarrow$ & Rank Corr. $\uparrow$ \\
         \midrule
         Independent Networks & \res{0.04}{0.00} & \res{0.99}{0.00} & \res{6.15}{0.61} & \res{0.82}{0.04} \\
         Shared Backbone with Two Heads & \res{0.04}{0.00} & \res{0.99}{0.00} & \res{7.31}{1.53} & \res{0.87}{0.02} \\
         \bottomrule
    \end{tabular}
\end{table}

\section{CIFAR-10H Experiment}

\subsection{Contrastive Learning on CIFAR}

To test whether the predicted certainty $\hat{\kappa}(x)$ aligns with human-judged aleatoric uncertainty, we require a dataset that provides a ground-truth. CIFAR-10H \cite{peterson2019human} provides 50 annotations for each test-set image of CIFAR-10. We use the entropy of the probability distribution over these annotations as a measure of aleatoric uncertainty in each image, and compare its negative to the predicted certainty $\hat{\kappa}(x)$ via rank correlation. Since the annotations were only collected for the 10000 images of the test set of CIFAR-10, we apply a 5-fold cross validation. The 10000 images are randomly split into sets of 2000. For five iterations, three of these sets form the train data, one the validation, and one the test data. To prevent confusions with the CIFAR-10 train and test set, we refer to these as the CIFAR-10H train, validation, and test sets. The image indices that belong to each set are provided in our code repository.

This leaves us with the task of redefining the CIFAR classification task into a contrastive learning problem. To this end, we simply assume that images are positive to one another if they belong to the same class and negative if they do not. CIFAR-10H, however, has soft class distributions for each image instead of a crisp class. Thus, we first draw a class $c$ from the class distribution $P(C|x)$ of a reference image $x$ from the train set. We then draw a positive image $x^+$ from a multinomial distribution over all train images weighed by their probabilities of that class $P(C=c|x^+)$. Negative images $x^-$ are selected the same way, but weighed by the probability of \emph{not} being class $c$, i.e., $1 - P(C=c|x^-)$. This provides the contrastive data generator required for training.

Since the human annotation data might be noisy in how well it captures the aleatoric uncertainty, we complement it with a synthetical way to introduce aleatoric uncertainty. In a second test dataset, we copy the CIFAR-10H test images, but perform a random crop and rescale that reduces the image to a proportion $\texttt{crop\_size} \sim \text{Unif}([0.25, 1])$ of its original width and length. This directly reduces the information available in the image and therefore increases its aleatoric uncertainty, without introducing artifacts that might let the image go out-of-distribution. We calculate the rank correlation of the reduction in size \texttt{crop\_size} and the (negative) predicted certainty $-\hat{\kappa}(x)$ as an alternative way to evaluate whether $\hat{\kappa}(x)$ reflects loss in information in the input, and therefore aleatoric uncertainty.

\subsection{Hyperparameters}

We use a ResNet-18 \cite{he2016deep} pretrained on the CIFAR-10 train dataset \cite{huy_phan_2021_4431043} and replace the classification layer by a linear layer with the input and output dimensions $[512, D]$. We then train the linear layer and the ResNet backbone under each loss for $8192$ batches of batchsize $128$, which corresponds to roughly 175 epochs on the $6000$ CIFAR-10H train images. We use the CIFAR-10H validation set to select the best model, evaluated after each 16 batches. The criterion is the rank correlation between $\hat{\kappa}(x)$ and the crop size in the synthetically deteriorated CIFAR-10H validation set. We chose this metric rather than the human annotator disagreement since it can be generated on arbitrary datasets without new annotations. All losses use $128$ MC samples and, according to the results in \cref{sec:hyp}, a fixed $\kappa_\text{pos}$. We use the same Adam optimizer with a learning rate of $0.0001$, learning rate scheduling, and (optional) phase-wise training as in \ref{sec:hyp}. The remaining hyperparameters were tuned via grid search. The best choices are highlighted in \cref{tab:hyperparams_cifar}.

\begin{table}[h]
    \centering
    \resizebox{\textwidth}{!}{
        \begin{tabular}{lccccc}
        \toprule
             Loss & HIB & ELK & MCInfoNCE & MCInfoNCE & MCInfoNCE\\
             Train Dataset / Label Type & CIFAR-10H soft & CIFAR-10H soft & CIFAR-10H soft & CIFAR-10H hard & CIFAR-10 hard \\
             \midrule
             Latent Dim $D$ & $\{\mathbf{8}, 16\}$ & $\{\mathbf{8}, 16\}$ & $\{\mathbf{8}, 16\}$ & $\{8,  \mathbf{16}\}$ & $\{\mathbf{8},  16\}$ \\
             Number of negatives $M$ & $\{\mathbf{0}, 1, 32\}$ & $\{0, \mathbf{1}, 32\}$&  $\{0, 1, \mathbf{32}\}$ & $\{\mathbf{0}, 1, 32\}$ & $\{\mathbf{0}, 1, 32\}$ \\
             $\kappa_\text{pos}$ & $\{16, \mathbf{32}, 64\}$ & $\{16, \mathbf{32}, 64\}$ & $\{\mathbf{16}, 32, 64\}$ & $\{\mathbf{16}, 32, 64\}$ & $\{16, 32, \mathbf{64}\}$\\
             Phasewise training & \{\textbf{yes}, no\}& \{yes, \textbf{no}\}& \{\textbf{yes}, no\} & \{yes, \textbf{no}\} & \{\textbf{yes}, no\}\\
             $a$ & $\{ 0.5, 1, \mathbf{2}, 4 \}$ \\
             $b$ & $\{-2, -1, 0, \mathbf{1}, 2\}$ \\
             \bottomrule
        \end{tabular}
    } 
    \caption{Possible hyperparameters and best-performing hyperparameters (\textbf{bold}). $M = 0$ corresponds to not sampling negatives, but using one sample from the same batch as a negative. HIB's additional hyperparameters were tuned after the first four parameters to reduce the number of grid-search evaluations.}
    \label{tab:hyperparams_cifar}
\end{table}

\subsection{Ablation without Pretraining}

All experiments on CIFAR started from weights pretrained on CIFAR-10 to reduce the required computational resources. However, it is also an intriguing question if MCInfoNCE is able to train a network from scratch. \cref{tab:cifar_pretraining} shows that it achieves a similar performance to when it is used on pretrained weights. The small gap in performance may be explained by the fact that we chose the same hyperparameters for both scenarios for fairness. In particular, the learning rate is tuned for the pretrained scenario but not for the non-pretrained one.

\begin{table}[h]
    \centering
    \caption{MCInfoNCE can also be used to train on CIFAR-10H from scratch, without pretraining. Rank correlation on unseen test data.}
    \label{tab:cifar_pretraining}
    \begin{tabular}{lcc}
    \toprule
         Pretraining & Annotator Entropy $\uparrow$ & Crop Size $\uparrow$ \\
         \midrule
         With Pretraining & 0.33 & 0.70 \\
         From Scratch & 0.31 & 0.62 \\
         \bottomrule
    \end{tabular}
\end{table}

\subsection{Uncertainty Estimation is Not At Stakes With First-Moment Estimation}
It is a popular question whether uncertainty estimation worsens the general performance, i.e., the estimation of the first-moment embedding $\hat{\mu}(x)$. To add evidence to this discussion, we've implemented the normal InfoNCE loss which estimates only $\hat{\mu}(x)$ but not $\hat{\kappa}(x)$. In both for the CIFAR and controlled experiment. \cref{tab:infonce} shows that MCInfoNCE is not worse than InfoNCE at predicting $\hat{\mu}(x)$. In terms of the RMSE in the controlled experiment, it even outperforms InfoNCE as InfoNCE puts the embeddings too close to one another (RMSE $=0.83$). This is although InfoNCE was hyperparameter-tuned.

\begin{table}[h]
    \centering
    \caption{MCInfoNCE is not worse than InfoNCE at predicting the first moment of the embedding despite also providing a variance estimate.}
    \label{tab:infonce}
    \begin{tabular}{lccc}
    \toprule
         Loss & $\mu(x)$ vs $\hat{\mu}(x)$ RMSE $\downarrow$ & $\mu(x)$ vs $\hat{\mu}(x)$ Rank Corr. $\uparrow$ & Recall@1 on CIFAR-10H $\uparrow$ \\
         \midrule
         MCInfoNCE & \res{0.04}{0.00} & \res{0.99}{0.00} & 0.863 \\
         InfoNCE & \res{0.83}{0.00} & \res{0.99}{0.00} & 0.858 \\
         \bottomrule
    \end{tabular}
\end{table}

\subsection{Credible Intervals}

Since we have a (estimated) posterior distribution $P(z|x)$, we can give a credible interval $\text{CI}_p \subseteq \mathcal{Z}$ that the latent $z$ of $x$ falls into with a probability $p \in [0, 1]$, i.e., $P(z \in \text{CI}_p) = p$. We center this interval around the mode of the posterior vMF, such that it is a highest posterior density interval (HPDI). Due to the rotational symmetry of the vMF, for a given $\kappa(x)$ and credible level $p$, this interval has the form $\text{CI}_p = \{z \in \mathcal{Z} | z^\top\mu(x) \leq t\}$, i.e., all latents $z$ closer to the mode $\mu(x)$ than a certain threshold $t \in [-1, 1]$ measured by cosine similarity. This threshold is the (approximated) $(1-p)$ quantile of the vMF.

To visualize this latent interval, we define the credible images interval (CII). This is a pre-image of the corresponding CI and gives all images whose mode is within the CI, i.e., $\text{CII}_p := \{x \in \mathcal{X} | \mu(x) \in \text{CI}_p \}$. This can either be visualized via a GAN conditional on $z \in \text{CI}_p$ or by images from the dataset with $\mu(x) \in \text{CII}_p$. We note that this does not reflect the aleatoric uncertainty of those images. We leave this extension for future work.  

\subsection{Qualitative Evaluation of Aleatoric Uncertainty}

\begin{figure}[h]
    \centering
    \begin{tikzpicture}
    \node[anchor=south west,inner sep=0] (image) at (0,0) {\includegraphics[trim=6.2cm 5.5cm 24.7cm 4cm, clip, width=0.33\textwidth]{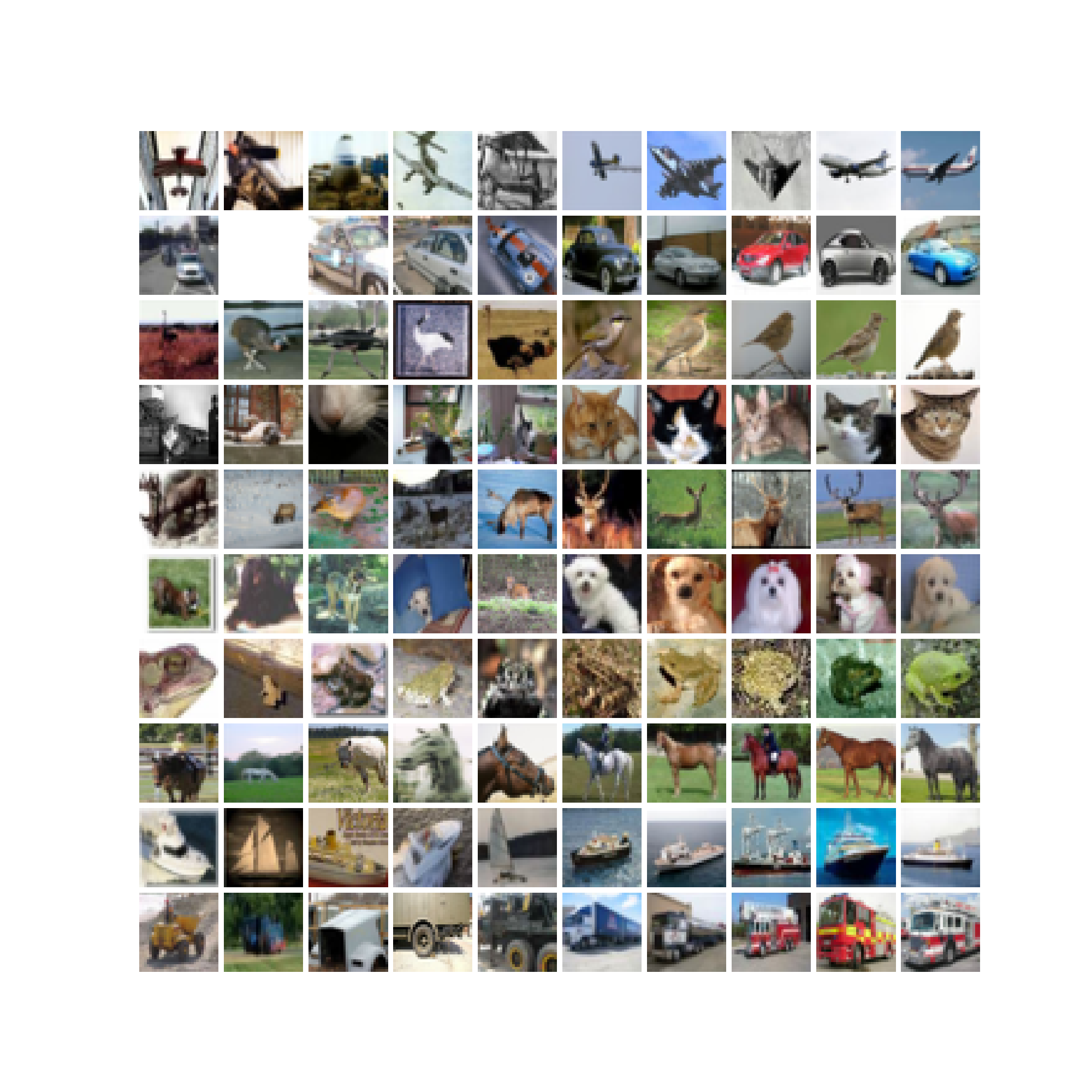}};
    \node[right=0.6cm of image,inner sep=0, align=center] (image0) {\includegraphics[trim=26.1cm 5.5cm 5.1cm 4cm, clip, width=0.33\textwidth]{figures/uncertain_images_testset.png}};
    \begin{scope}[x={(image.south east)},y={(image.north west)}]
        \node [anchor=south] at (0.5,0.95) {\footnotesize Lowest $\hat{\kappa}(x)$};
        \node [anchor=south] at (1.61,0.95) {\footnotesize Highest $\hat{\kappa}(x)$};
        
        \node [anchor=west] at (-0.45,0.05) {\footnotesize Class=Truck};
        \node [anchor=west] at (-0.45,0.144) {\footnotesize Class=Ship};
        \node [anchor=west] at (-0.45,0.239) {\footnotesize Class=Horse};
        \node [anchor=west] at (-0.45,0.333) {\footnotesize Class=Frog};
        \node [anchor=west] at (-0.45,0.43) {\footnotesize Class=Dog};
        \node [anchor=west] at (-0.45,0.522) {\footnotesize Class=Deer};
        \node [anchor=west] at (-0.45,0.617) {\footnotesize Class=Cat};
        \node [anchor=west] at (-0.45,0.711) {\footnotesize Class=Bird};
        \node [anchor=west] at (-0.45,0.806) {\footnotesize Class=Automobile};
        \node [anchor=west] at (-0.45,0.9) {\footnotesize Class=Airplane};
    \end{scope}
    \end{tikzpicture}
    \caption{Images for which MCInfoNCE predicts the highest aleatoric uncertainty , i.e., lowest $\hat{\kappa}(x)$, (left) per class qualitatively look more ambiguous than those with the highest predicted $\hat{\kappa}(x)$ (right).}
    \label{fig:unc_images}
\end{figure}

Besides the quantitative metrics reported in the main text, we can also take a qualitative look at whether $\hat{\kappa}(x)$ represents aleatoric uncertainty in the inputs. \cref{fig:unc_images} visualizes the five images with the lowest and highest $\hat{\kappa}(x)$ in each class in the CIFAR-10H test set, i.e., on unseen data. It can be seen that images with a low $\hat{\kappa}(x)$ tend to hide characteristic parts of the object via bad crops, being too far away from the object, or an uncommon perspective. Images with a high $\hat{\kappa}(x)$ show characteristic features clearly, making it less ambiguous to tell what they show. In other words, they indeed have a lower aleatoric uncertainty.

\end{document}